\documentclass[preprint, 3p,authoryear]{elsarticle}

\journal{International Journal of Greenhouse Gas Control}

\usepackage{amsmath}
\usepackage{amsfonts}
\usepackage{graphicx}
\usepackage{amsthm}
\usepackage{moreverb}
\usepackage{amsfonts}
\usepackage{mathrsfs}
\usepackage{float}
\usepackage{fixltx2e}
\usepackage{gensymb}
\usepackage{xfrac}
\usepackage{lmodern}
\usepackage{color}
\usepackage{caption}
\usepackage{subcaption}
\usepackage{epstopdf}
\usepackage{listings}
\usepackage{todonotes}
\usepackage{booktabs}
\usepackage{multirow}
\usepackage{array}
\usepackage{tabularx}
\usepackage{placeins}
\usepackage{natbib}
\usepackage{appendix}
\usepackage{longtable}

\PassOptionsToPackage{hyphens}{url} 
\usepackage[hidelinks]{hyperref}    

\usepackage[ruled, lined, linesnumbered, commentsnumbered, longend]{algorithm2e}

\SetKwInOut{KwIn}{Input}
\SetKwInOut{KwOut}{Output}

\begin{document}

\begin{frontmatter}

\title{Integrating Score-Based Diffusion Models with Machine Learning-Enhanced Localization for Advanced Data Assimilation in Geological Carbon Storage}

\author[1,2]{Gabriel Serr{\~a}o Seabra \corref{mycorrespondingauthor}}
\cortext[mycorrespondingauthor]{Corresponding author}
\ead{g.serraoseabra@tudelft.nl}

\author[3]{Nikolaj T. M{\"u}cke}
\ead{nikolaj.mucke@cwi.nl}

\author[2,5]{Vinicius Luiz Santos Silva}
\ead{v.santos-silva19@alumni.imperial.ac.uk}

\author[2]{Alexandre A. Emerick}
\ead{alexandre.emerick@petrobras.com.br}

\author[1,6]{Denis Voskov}
\ead{voskov@tudelft.nl}

\author[1]{Femke Vossepoel}
\ead{f.c.vossepoel@tudelft.nl}

\address[1]{Faculty of Civil Engineering and Geosciences, TU Delft, Stevinweg 1,
2628 CN Delft, Netherlands}
\address[2]{Petroleo Brasileiro S.A. (Petrobras), Rio de Janeiro, Brazil}
\address[3]{Centrum Wiskunde \& Informatica, Science Park 123, 1098 XG Amsterdam, Netherlands}
\address[5]{Novel Reservoir Modelling and Simulation Group, Imperial College London, London, United Kingdom}
\address[6]{Department of Energy Resources Engineering, Stanford University, CA, USA}

\begin{abstract}
Accurate characterization of subsurface heterogeneity is important for the safe and effective implementation of geological carbon storage (GCS) projects. This paper explores how machine learning methods can enhance data assimilation for GCS with a framework that integrates score-based diffusion models with machine learning-enhanced localization in channelized reservoirs during CO$_2$ injection. We employ a machine learning-enhanced localization framework that uses large ensembles ($N_s = 5000$) with permeabilities generated by the diffusion model and states computed by simple ML algorithms to improve covariance estimation for the Ensemble Smoother with Multiple Data Assimilation (ESMDA). We apply ML algorithms to a prior ensemble of channelized permeability fields, generated with the geostatistical model FLUVSIM. Our approach is applied on a CO$_2$ injection scenario simulated using the Delft Advanced Research Terra Simulator (DARTS). Our ML-based localization maintains significantly more ensemble variance than when localization is not applied, while achieving comparable data-matching quality. This framework has practical implications for GCS projects, helping improve the reliability of uncertainty quantification for risk assessment.
\end{abstract}


\end{frontmatter}


\section{Introduction}\label{section:introduction}

Geological carbon storage (GCS) is a promising technology for achieving large-scale carbon dioxide removal from the atmosphere, which is required to meet global climate targets \citep{bui2018carbon}. The Intergovernmental Panel on Climate Change (IPCC) estimates that between 350 and 1,200 gigatons of CO$_2$ must be captured and stored by 2100 to limit global warming to 1.5°C above pre-industrial levels. This unprecedented scale demands advanced methodologies for characterizing and monitoring subsurface storage complexes to ensure safe and permanent sequestration while maximizing storage efficiency \citep{ringrose2021store}. Data assimilation provides the mechanism to combine prior geological models with monitoring data to reduce uncertainty and improve forecasts, but practical deployments are constrained by computational resources as each assimilation cycle requires many forward simulations. This motivate adding machine learning components that assist DA without replacing the governing physics—for example by generating large prior ensembles and improving localization for small $N_e$. Long-term experience from Sleipner \citep{furre2017co2} and established best practices \citep{chadwick2008best} inform the monitoring context in which we develop and test the approach.

Wells typically sample a small part of the reservoir volume, yet decisions affecting containment must be made based on this limited information. This data scarcity becomes particularly challenging in channelized formations, which can be attractive storage targets because of their high injectivity, where complex connectivity patterns defy simple statistical descriptions \citep{pyrcz2014geostatistical}.

Existing geostatistical methods face inherent limitations when confronted with channelized systems. Two-point statistics cannot capture curvilinear channel geometries, whereas multiple-point statistics require training images that may poorly represent the true subsurface heterogeneity \citep{mariethoz2010multiple,strebelle2002conditional}. When these prior models are updated through ensemble-based data assimilation, the limited number of ensemble members exacerbates these problems. Small ensemble sizes (typically $N_e \sim 50-200$ in operational settings) lead to poorly estimated covariance matrices with significant sampling errors, causing the assimilation update to introduce geologically implausible features, such as spurious high-permeability patches. In practice, $N_e$ is capped by simulation cost per member, parallel scalability limits, and project turnaround requirements. The insufficient ensemble size worsens the results in artificial blurring of sharp geological boundaries, as the linear update mechanism cannot preserve the discrete nature of channel-background contrasts. This geological degradation becomes more severe with smaller ensembles, where sampling errors dominate the true covariance structure, ultimately producing posterior realizations that bear little resemblance to the complex channelized patterns that control the subsurface flow behavior. Although object-based geological models, such as FLUVSIM \citep{deutsch1992fluvsim}, can generate geologically realistic channelized systems that address these representational limitations, they are computationally expensive and time-consuming to run, making it impractical to generate the hundreds or thousands of realizations required on-the-fly during data assimilation workflows.

Data assimilation methods provide the mathematical framework for combining these prior geological models with observed data to reduce uncertainty in reservoir characterization \citep{Oliver2008, evensenDataAssimilationFundamentals2022}. The fundamental approaches can be broadly categorized into variational methods, which solve optimization problems to find the most probable model realization \citep{tarantolaInverseProblemTheory2005}, and ensemble methods, which use an ensemble to approximate probability distributions \citep{evensen2009data}. Ensemble methods quantify uncertainty through their sample-based representation, making them suitable for GCS applications where understanding the range of possible outcomes is as important as identifying the most likely scenario \citep{aanonsen2009ensemble}. The evolution from the original Ensemble Kalman Filter \citep{evensen1994sequential} to smoothers and iterative variants reflects ongoing efforts to handle increasingly complex nonlinear problems while maintaining computational feasibility \citep{emerick2013esmda,chen2012ermles}. Advances address the challenge of limited ensemble sizes through localization techniques \citep{houtekamer2016review} and iterative schemes that linearize nonlinear problems \citep{oliver2011recent}. However, these methods still face challenges when applied to highly heterogeneous channelized systems, where traditional assumptions about spatial correlation and Gaussian distributions break down.

The Ensemble Smoother with Multiple Data Assimilation (ESMDA) is a common method of choice for practical reservoir characterization due to its computational efficiency and ability to handle nonlinear forward models \citep{emerick2013esmda,evensen2009data}. However, ESMDA has Gaussian assumptions that channelized permeability fields severely violate, and this can exacerbate the fact that small ensembles lose their variance, causing posterior models to be less uncertain than they should be, which affects the reliability of uncertainty quantification.  The industry has widely adopted distance-based localization methods \citep{hamill2001distance,gaspari1999construction,houtekamer2016review} as the primary solution to combat ensemble collapse. Covariance estimation in high-dimensional systems \citep{furrer2007estimation} and distance based methods, including the popular Gaspari-Cohn correlation function, are standard practice in several operational reservoir history matching systems. While those approaches suppress spurious correlations arising from sampling errors, they rely on an assumption that correlation strength decreases with physical distance — an assumption that fails in channelized systems. A circular or elliptical localization kernel cannot distinguish between updating along a high-permeability channel versus across a low-permeability barrier, leading to geologically inconsistent modifications that compromise long-term predictions. Despite these limitations, distance-based localization remains the industry standard due to its computational efficiency and ease of implementation. Recent work has shown the benefits of applying covariance localization to field cases \citep{emerick2011history} and comparing different adaptive localization methods for reservoir characterization \citep{lacerda2019comparison}.

Score-based diffusion models, generative AI models that learn to approximate the gradient of the log-density ($\nabla_x \log p_t(x)$) and can iteratively denoise random noise into realistic samples \citep{song2021score,song2019generative}, offer a promising approach for ensemble generation by training on established geological modeling outputs, such as those from FLUVSIM \citep{deutsch1992fluvsim}, to capture the complex spatial patterns and connectivity structures inherent in channelized systems. This capability addresses a need in ensemble-based data assimilation, where accurate covariance estimation benefits from much larger sample sizes than typically computationally feasible \citep{furrer2007estimation,houtekamer2001sequential}. Once trained, diffusion models can rapidly generate large numbers of geologically consistent realizations on demand, providing super-ensembles for enhanced covariance estimation without requiring additional, expensive reservoir simulations.  The generated realizations preserve essential geological features from the training data, including channel geometries and connectivity patterns that control fluid flow behavior in complex heterogeneous formations \citep{pyrcz2014geostatistical}. By leveraging these large synthetic ensembles within machine learning-enhanced localization frameworks \citep{silva2025mitigating,silva2025machine,lacerda2021machine}, this approach can improve covariance estimation accuracy while maintaining the geological realism necessary for reliable uncertainty quantification in carbon storage applications.

Recent advances have demonstrated the potential of integrating machine learning with data assimilation across various fields. The evolution of DA-ML integration has progressed from early demonstrations on low-dimensional chaotic systems \citep{brajard2020combining} to frameworks that merge DA and ML through unified Bayesian formulations \citep{bocquet2020bayesian}. The designers of these frameworks established that DA can provide complete state reconstructions from sparse observations, enabling ML models to learn dynamics that would otherwise be untrainable. The concept of "Data Learning"—combining ML's pattern recognition capabilities with DA's physics-based constraints—has emerged as a powerful paradigm \citep{Buizza2022}, with theoretical foundations showing how many DA-ML approaches can be unified under expectation-maximization frameworks.

In the subsurface domain, researchers explored two primary integration strategies: using ML to enhance DA processes and leveraging DA/UQ methods to improve ML models through physics-informed constraints \citep{Cheng2023}. For enhancing DA, ML is applied to learn model error corrections \citep{farchi2021using,farchi2021comparison}, infer unresolved scale parametrizations directly from observations \citep{brajard2021combining}, and enable training with partial observations through hybrid forecasting approaches \citep{wikner2021using}. Recent theoretical advances include Deep Data Assimilation \citep{arcucci2021deep}, where neural networks learn to perform assimilation updates internally, embedding DA knowledge into the model itself. For geological carbon storage specifically, the integration of generative AI with DA represents a paradigm shift. Recent work demonstrates that latent diffusion models can parameterize complex facies-based geomodels while maintaining geological realism during history matching \citep{difederico2024latent}, working in latent space through VAE (Variational Auto Encoders) encoders. Score-based diffusion models have enabled real-time ensemble forecasts of 3D CO$_2$ plume evolution, generating hundreds of realizations in minutes rather than hours \citep{fan2024conditional}, although they require extensive training datasets. Beyond diffusion models, researchers employ conditional normalizing flows within Bayesian filtering frameworks to create uncertainty-aware "digital shadows" for continuous monitoring \citep{gahlot2025uncertainty}. They have also developed hierarchical DA approaches using sequential Monte Carlo methods to handle uncertainty in both probability distributions and structural assumptions \citep{teng2025likelihood}. While these advances demonstrate that generative models can capture non-Gaussian uncertainties and maintain geological consistency, the challenge of adaptive localization that respects complex geological structures in ensemble-based methods remains an active area of research.

To explore how machine learning can enhance data assimilation for GCS, we combine score-based diffusion models with machine learning-enhanced localization to overcome these limitations. Our investigation demonstrates that ML methods can improve data assimilation performance through an approach that operates on three levels:

\begin{enumerate}
\item \textbf{Efficient super-ensemble generation}: We employ score-based diffusion models trained on geostatistical realizations to enable the generation of thousands of permeability field samples. This overcomes the computational bottleneck of traditional methods by providing large ensembles (5000 members) for statistical estimation without running expensive forward simulations.

\item \textbf{Enhanced covariance estimation via ML proxies}: By training fast surrogate models on the working ensemble and applying them to large super-ensembles (5000 members) from the diffusion model, we improve covariance estimation accuracy without additional forward simulations.

\item \textbf{Adaptive localization respecting geology}: The ML-enhanced covariances capture how information propagates through specific geological structures, enabling localization that adapts to channel connectivity rather than imposing arbitrary geometric constraints.
\end{enumerate}

We test our method on the methodology for 2D channelized systems with pressure observations. We do not address real-time monitoring, dynamic model updating, or non-channelized geological settings—each worthy of dedicated investigation building upon this foundational work.

\section{Theoretical Background}\label{section:theory}

This section presents the theoretical foundations of our integrated framework, covering score-based diffusion models, ensemble data assimilation with localization, and machine learning-enhanced covariance estimation.

\subsection{Score-Based Diffusion Models: Mathematical Foundation}

\begin{figure}[htbp]
  \centering
  \includegraphics[width=\textwidth]{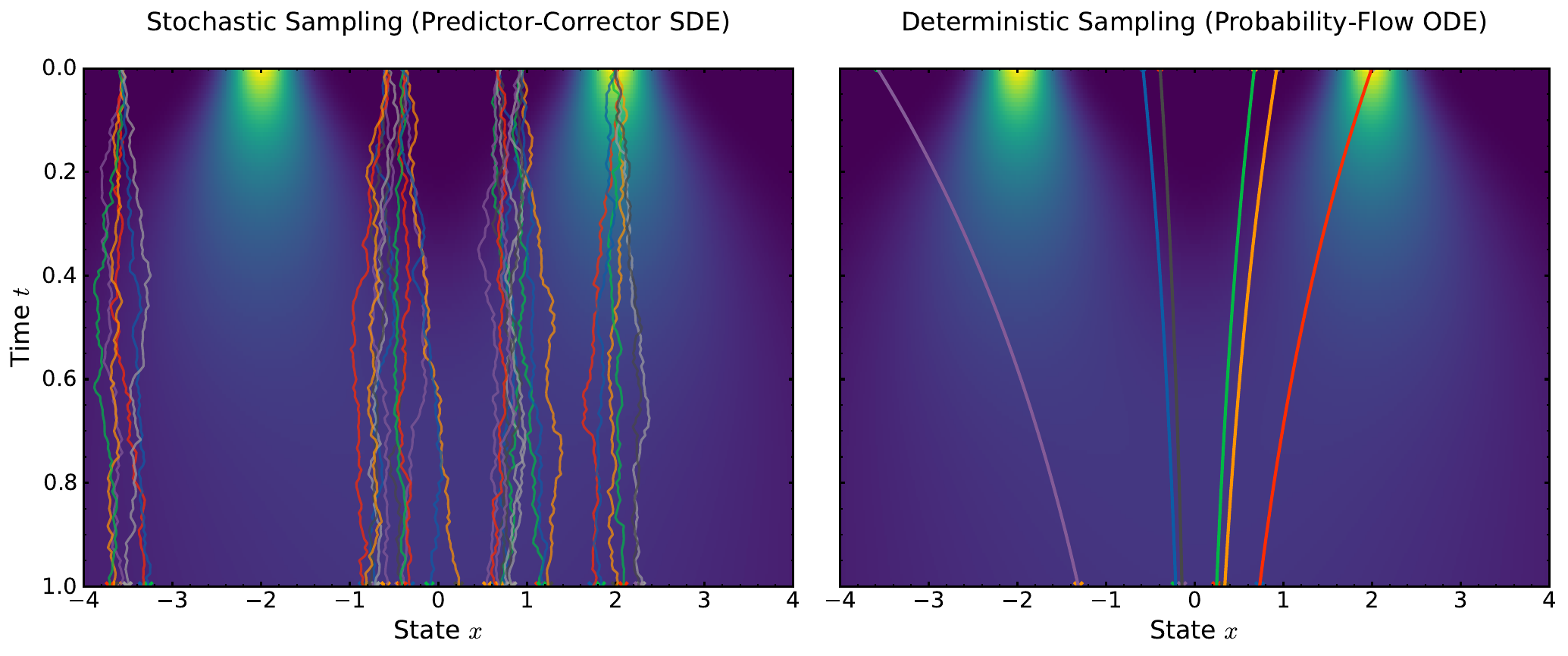}
  \caption{Score-based diffusion process. Forward diffusion (top) transforms channelized permeability to noise via VE-SDE. Reverse process (bottom) uses learned score function $s_\theta(\mathbf{x}_t, t)$ for denoising with SDE (stochastic) or ODE (deterministic) sampling.}
  \label{fig:score_diffusion_theory}
  \end{figure}

Score-based diffusion models add noise to data through a forward process and then learn to reverse it to generate new samples \citep{song2019generative,song2021score}. These methods evolved from foundational work on deep unsupervised learning using nonequilibrium thermodynamics \citep{sohl2015deep} and denoising diffusion probabilistic models \citep{ho2020denoising}, and have demonstrated better performance when compared to GANs (Generative Adversarial Networks) and variational autoencoders for high-quality sample generation \citep{dhariwal2021diffusion}. Figure~\ref{fig:score_diffusion_theory} illustrates the complete forward and reverse diffusion process, showing how channelized permeability fields are progressively transformed into noise and then reconstructed. The forward process transforms data samples into random noise through a continuous-time stochastic process defined by a stochastic differential equation (SDE):
\begin{equation} \label{eq:forward_sde}
\mathrm{d} \mathbf{x}_t = g(t) \mathrm{d} \mathbf{W}_t, \quad t\in[0,T]
\end{equation}
where the initial condition, $\mathbf{x}_0 \in \mathbb{R}^{D}$, is a sample from the data distribution representing the permeability field in our context. The diffusion coefficient $g(t)$ controls the noise injection rate, $\mathbf{W}_t$ is the Wiener process (Brownian motion), and $t \in [0, T]$ is the time variable. As time progresses from $t=0$ to $t=T$, $\mathbf{x}_t$ gradually transitions from a data distribution to a noise distribution.

In this work, we adopt the Score Matching with Langevin Dynamics (SMLD) variant of the Variance Exploding (VE) SDE formulation \citep{song2019generative}. In this framework, the forward diffusion is defined as $d\mathbf{x} = \sigma^t d\mathbf{W}$, where $\sigma = 25.0$ is the noise scale parameter. The diffusion coefficient is given by $g(t) = \sigma^t$, and the marginal probability standard deviation is:
\begin{equation} \label{eq:sigma}
\sigma_t = \sqrt{\frac{\sigma^{2t} - 1}{2\ln\sigma}}
\end{equation}
This formulation ensures that the variance of the noisy data grows exponentially with time, providing a transition from structured channelized patterns to isotropic Gaussian noise. The choice of $\sigma = 25.0$ was empirically determined to provide sufficient noise levels to destroy geological structures while maintaining numerical stability during training. The conditional distribution $p_{t}(\mathbf{x}_t | \mathbf{x}_0)$ follows a Gaussian distribution:
\begin{equation} \label{eq:diffusion_conditional_distribution}
p_{t}(\mathbf{x}_t | \mathbf{x}_0) = \mathcal{N}(\mathbf{x}_0, \sigma^2_t\mathbf{I})
\end{equation}
The key insight is that the time-reversed SDE can generate samples from the data distribution:
\begin{equation} \label{eq:reverse_sde}
\mathrm{d}\mathbf{x}_t = - g^2(t)\nabla_{\mathbf{x}_t}\log p_t(\mathbf{x}_t)\mathrm{d} t + g(t)\mathrm{d}\bar{\mathbf{W}}_t, \quad t\in[T,0]
\end{equation}
where $\nabla_{\mathbf{x}_t}\log p_t(\mathbf{x}_t)$ is the score function — the gradient of the log-probability density. By solving this reverse SDE from $t=T$ to $t=0$ with initial condition $\mathbf{x}_T \sim \mathcal{N}(0, \sigma_T^2\mathbf{I})$, we obtain samples from the data distribution.

The score function is approximated using a neural network $s_\theta(\mathbf{x}_t, t) \approx \nabla_{\mathbf{x}_t}\log p_t(\mathbf{x}_t)$, trained using denoising score matching \citep{vincent2011connection}:
\begin{equation} \label{eq:dsm_loss}
\mathcal{L}(\theta) = \mathbb{E}_{t \sim \mathcal{U}(0,T)} \mathbb{E}_{\mathbf{x}_0 \sim p_\text{data}} \mathbb{E}_{\mathbf{x}_t \sim p_{t}(\mathbf{x}_t | \mathbf{x}_0)} \left\| s_\theta(\mathbf{x}_t, t) - \nabla_{\mathbf{x}_t} \log p_{t}(\mathbf{x}_t | \mathbf{x}_0) \right\|^2
\end{equation}
The trained diffusion model can also be adapted for posterior sampling to condition generated realizations on direct permeability observations. This approach follows recent advances in diffusion posterior sampling for general noisy inverse problems \citep{chung2023diffusion}, which demonstrated how diffusion models can be adapted for Bayesian inference tasks. This capability, while not used in the main data assimilation workflow, provides an additional tool for integrating hard data constraints. We present the theoretical foundation and validation of this posterior sampling approach in \ref{appendix:posterior}.

The reverse diffusion process in Equation \ref{eq:reverse_sde} can be solved using two distinct approaches: stochastic (SDE) or deterministic (ODE) sampling. The probability flow ODE, which shares the same marginal distributions as the SDE, is given by:
\begin{equation} \label{eq:probability_flow_ode}
\frac{\mathrm{d}\mathbf{x}_t}{\mathrm{d}t} = -\frac{1}{2}g^2(t)\nabla_{\mathbf{x}_t}\log p_t(\mathbf{x}_t), \quad t\in[T,0]
\end{equation}
We tested both SDE and ODE formulations for ensemble generation in data assimilation, and both approaches yield good results for generating geologically realistic channelized permeability fields. The choice between methods depends on the specific application requirements: SDE sampling provides more stochastic variability and exploration of the learned distribution, while ODE sampling offers greater control and reproducibility. For our application in ensemble generation for data assimilation, we adopt the ODE formulation for several practical advantages. The deterministic nature ensures reproducibility. Given its deterministic nature and a fixed starting noise, the ODE always produces the same realization, enabling controlled experiments and sensitivity analyses. The ODE formulation typically requires fewer discretization steps than the SDE for comparable quality, improving computational efficiency. The smooth trajectories of the ODE solver also facilitate better preservation of geological structures during the denoising process. However, both the reverse-time SDE and the probability-flow ODE can generate large ensembles of geologically realistic samples suitable for enhanced covariance estimation in our setting, since they both are able to generate realistic samples.

\subsection{Ensemble-Based Data Assimilation and Localization}

The Ensemble Smoother with Multiple Data Assimilation (ESMDA) \citep{emerick2013esmda} provides a framework for integrating dynamic observations into reservoir models. Extending fundamental ensemble Kalman filter theory \citep{evensen1994sequential,evensen2009data} as described in the unified data assimilation framework of \citet{evensenDataAssimilationFundamentals2022}, the method performs multiple assimilation steps with \emph{inflated} observation-error covariances, thereby linearizing the nonlinear inverse problem through iteration. Here, inflation means that the observation-error covariance matrix $\mathbf{C}_D$ is multiplied by a scalar factor $\alpha_i>1$ at each step, and the corresponding random perturbations of the data are scaled by $\sqrt{\alpha_i}$. This distributes a single Bayesian update over several smaller ones, which improves linearization and reduces sampling noise.

For an ensemble of model parameters $\{\mathbf{z}_j\}_{j=1}^{N_e}$, where $\mathbf{z} \in \mathbb{R}^{N_z}$ represents the log-permeability field, the ESMDA update equation at iteration $i$ is:
\begin{equation} \label{eq:esmda_update}
\mathbf{z}_j^{a} = \mathbf{z}_j^{f} + \mathbf{C}_{zD}^{f} \left( \mathbf{C}_{DD}^{f} + \alpha_i \mathbf{C}_D \right)^{-1} \left( \mathbf{d}_{\text{obs}} + \sqrt{\alpha_i} \boldsymbol{\epsilon}_j - \mathbf{d}_j^{f} \right)
\end{equation}
where superscripts $f$ and $a$ denote forecast (prior) and analysis (posterior) quantities respectively, $\mathbf{C}_{zD}^{f}$ is the cross-covariance between parameters and predicted data, $\mathbf{C}_{DD}^{f}$ is the auto-covariance of predicted data, $\mathbf{C}_D$ is the observation error covariance matrix, $\alpha_i$ is the inflation coefficient for iteration $i$, $\mathbf{d}_{\text{obs}}$ represents the observed data, $\mathbf{d}_j^{f} = \mathbf{G}(\mathbf{z}_j^{f})$ is the predicted data from forward model $\mathbf{G}$, and $\boldsymbol{\epsilon}_j \sim \mathcal{N}(0, \mathbf{C}_D)$ are random perturbations.

However, for practical ensemble sizes (typically $N_e \sim 100$), the sample covariance matrices contain large errors, leading to spurious correlations and subsequent ensemble collapse \citep{houtekamer2001sequential}. The industry standard solution to deal with these spurious correlations is distance-based localization, which modifies the Kalman gain through element-wise multiplication with a localization matrix $\mathbf{L}$. This matrix typically uses correlation functions such as the fifth-order piecewise rational function of Gaspari and Cohn \citep{gaspari1999construction}, which tapers correlations to zero beyond a specified distance, in other words, it masks part of the error covariance matrix:
\begin{equation}
\mathbf{z}_j^{a} = \mathbf{z}_j^{f} + \mathbf{L} \circ \left[ \mathbf{C}_{zD}^{f} \left( \mathbf{C}_{DD}^{f} + \alpha_i \mathbf{C}_D \right)^{-1} \right] \left( \mathbf{d}_{\text{obs}} + \sqrt{\alpha_i} \boldsymbol{\epsilon}_j - \mathbf{d}_j^{f} \right)
\end{equation}
where $\circ$ denotes the Schur (element-wise) product.

Pseudo-optimal localization (PO), as we use it in the present study, denotes a taper derived by minimizing the expected Frobenius-norm error between the true and localized covariance matrices, following the formulation of \cite{furrer2007estimation}. This yields data-driven coefficients that balance the removal of spurious correlations against the preservation of true correlations.

\subsection{Machine Learning-Enhanced Localization}

Distance-based localization methods have become the industry standard for ensemble-based data assimilation, with implementations ranging from simple cutoff functions, like Gaspari-Cohn \citep{gaspari1999construction}, to adaptive correlation-based localization models \cite{Fvossepoel2025adaptive} and Schur product-based approaches \citep{houtekamer2001sequential}. These methods have been successfully deployed in operational weather forecasting, oceanography, and petroleum reservoir history matching for over two decades. However, while effective for many applications, they fail to capture the complex, non-local relationships between channelized permeability structures and pressure observations. In channelized systems, where pressure information travels along high-permeability pathways over long distances, distance-only tapering can be misleading, while correlation-based schemes can follow the geology, provided the estimated correlations are reliable. The challenge is that accurate correlation estimates typically require large ensembles, which are rarely affordable in practice. To address this, we use machine learning to augment classical localization: a fast proxy is trained on the working ensemble and then applied to a large super-ensemble generated by the diffusion model to produce more stable, correlation-aware covariance estimates even when the operational ensemble $N_e$ is small \citep{lacerda2021machine,silva2025mitigating,silva2025machine}.

Our ML-enhanced localization framework extends the approach proposed by \citet{silva2025mitigating,silva2025machine} by training a fast proxy model $f^{ML}: \mathbf{z} \rightarrow \mathbf{d}$ using the working ensemble, then applying this proxy to a large super-ensemble, of around five thousand members, generated by the diffusion model. The enhanced covariance estimate becomes:
\begin{equation} \label{eq:ml_covariance_detailed}
\tilde{\mathbf{C}}_{zD}^{ML} = \frac{1}{N_s-1} \sum_{k=1}^{N_s} (\mathbf{z}_k^s - \bar{\mathbf{z}}^s)(f^{ML}(\mathbf{z}_k^s) - \bar{\mathbf{d}}^{ML})^T
\end{equation}
where $N_s \gg N_e$ is the super-ensemble size, and $\mathbf{z}_k^s$ are samples from the diffusion model. This approach increases the sample size for covariance estimation while maintaining computational feasibility.

Localization is effective when the taper $\mathbf{L}$ keeps parameter--data entries that represent true physical influence and zeros those that are spurious. With small ensembles, the sample cross–covariance $\mathbf{C}_{zD}$ is noisy: Monte Carlo error scales like $O(1/\sqrt{N_e})$, so many weak-but-real correlations are drowned out by sampling noise. Our approach trains a fast proxy $f^{ML}:\mathbf{z}\mapsto\mathbf{d}$ on the working ensemble and then evaluates it on a large super-ensemble ($N_s \gg N_e$) drawn from the diffusion model.

\section{Methods and Implementation}\label{section:methodology}

This section presents the technical details of our integrated framework, including the channelized reservoir model, diffusion model implementation, ESMDA configuration, and machine learning-enhanced localization approach.

\subsection{Channelized Reservoir Generation and CO$_2$ Injection Modeling}

We employ a 2D channelized reservoir model representative of fluvial depositional environments commonly encountered in CO$_2$ storage formations. The model domain consists of a 64×64 grid with 5-meter cell dimensions, resulting in a 320×320 meter area. Channelized permeability fields are generated using the FLUVSIM algorithm \citep{deutsch1992fluvsim}, which simulates fluvial channel systems through object-based modeling.

\begin{figure}[htbp]
\centering
\includegraphics[width=0.9\textwidth]{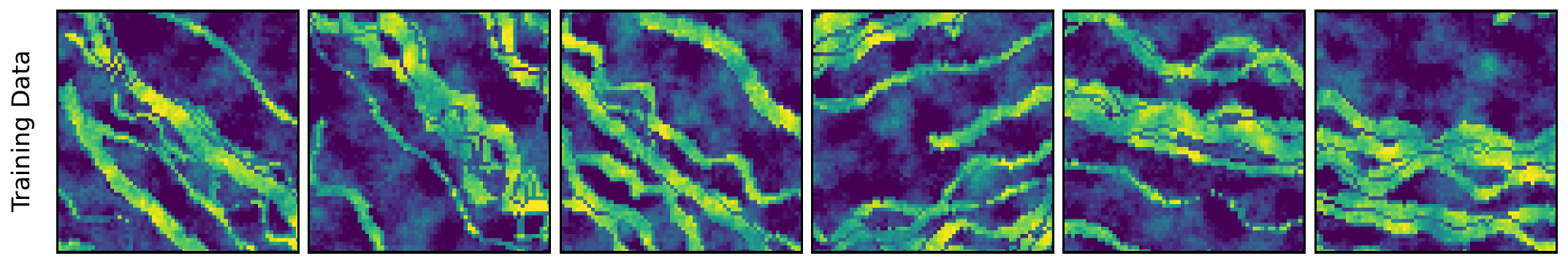}
\caption{Representative FLUVSIM-generated channelized permeability fields with bimodal distribution: channels (2000 mD) in background (50 mD).}
\label{fig:training_samples}
\end{figure}

The FLUVSIM parameters are sampled from normal distributions $\mathcal{N}(\mu, \sigma)$ to ensure geological diversity across the training dataset. Table~\ref{tab:fluvsim_params} summarizes the key parameters with their mean ($\mu$) and standard deviation ($\sigma$) values.
\begin{table}[htbp]
\centering
\caption{FLUVSIM parameters for channelized reservoir generation}
\label{tab:fluvsim_params}
\begin{tabular}{lll}
\hline
\textbf{Parameter} & \textbf{Mean $\pm$ Std. Dev.} & \textbf{Description} \\
\hline
Channel orientation (degrees) & $90 \pm 30$ & Primary flow direction \\
Channel amplitude (m) & $250 \pm 10$ & Variable sinuosity \\
Channel wavelength (m) & $2000 \pm 50$ & Meandering patterns \\
Width-to-thickness ratio &  $50 \pm 5$ & Channel aspect ratio \\
Channel proportion & $0.40 \pm 0.05$ & Net-to-gross ratio \\
Undulation wavelength (m) & $250 \pm 10$ & Thickness/width variations \\
\hline
\end{tabular}
\end{table}
This stochastic parameterization generates diverse channelized patterns while maintaining geological plausibility. The resulting permeability fields exhibit bimodal distributions with high-permeability channels (2000 mD) embedded in low-permeability background facies (50 mD), creating preferential flow paths characteristic of fluvial deposits. Figure~\ref{fig:training_samples} shows representative examples of the channelized permeability fields generated using FLUVSIM.

The CO$_2$ injection and storage simulations are performed using the Delft Advanced Research Terra Simulator (DARTS), a compositional flow simulator specifically developed for subsurface flow applications \citep{Voskov2017275}. DARTS uses operator-based linearization for efficient solution of complex multiphase flow problems \citep{Khait2017}, and has been validated against benchmark cases including the FluidFlower experiment \citep{wapperom2023fluidflower}. The simulator has been successfully applied to optimize CO$_2$ injection using multi-scale reconstruction \citep{chen2019optimization} and to model enhanced CO$_2$ dissolution trapping in saline aquifers \citep{Lyu2023}, demonstrating its capabilities for complex CO$_2$ storage scenarios. The reservoir model employs a structured Cartesian grid with dimensions of 64×64×1 cells, representing a two-dimensional horizontal layer. Each grid cell measures 5 meters in length and width, with a thickness of 10 meters, resulting in a computational domain of 320×320 meters horizontally. The simulated reservoir layer is situated at a depth of 2,000 meters, reflecting typical geological conditions for CO$_2$ storage, with an initial pressure of 150 bar and temperature of 50°C.

The reservoir simulations involve a two-component, two-phase system comprising carbon dioxide (CO$_2$) and water (brine with 100,000 ppm salinity). The permeability field is generated using FLUVSIM as previously described, exhibiting bimodal distribution with high-permeability channels (2000 mD) embedded in low-permeability background (50 mD). The porosity remains constant throughout the reservoir at 0.3.

CO$_2$ is injected through a single vertical well located centrally within the grid at a consistent rate of 1000 m$^3$/year over a two-year period. The injected fluid consists of nearly pure CO$_2$ into the saline aquifer. Pressure measurements are recorded monthly at four monitoring wells positioned in a five-spot pattern, each located at equal distances from the injection point in the cardinal directions. These observations provide spatial data points for subsequent data assimilation, with measurement errors of 1\% reflecting realistic field monitoring uncertainties.

The relative permeability in the porous medium is modeled using the Brooks-Corey formulation \citep{brooks1964hydraulic}, with exponents $n_w = 2.0$ for water and $n_g = 1.5$ for CO$_2$. Residual saturations are set to $S_{wc} = 0.25$ and $S_{gc} = 0.1$, with end-point relative permeability values of $k_{rw} = k_{rg} = 1.0$. Capillary pressure is neglected due to its limited influence within the two-dimensional modeling framework. Diffusion and hysteresis effects, while potentially influential on CO$_2$ migration and trapping, are excluded to maintain consistency with the study's focus on large-scale reservoir dynamics. This modeling approach is considered sufficient to capture the key multiphase flow mechanisms governing CO$_2$ migration and entrapment in porous media.

DARTS uses operator-based linearization for efficient solution of the coupled multiphase flow and transport equations, accurately capturing the complex physics of CO$_2$ injection including phase behavior and gravitational segregation. The simulator's computational efficiency enables the thousands of forward model evaluations required for ensemble-based data assimilation.

\begin{figure}[htbp]
\centering
\includegraphics[width=\textwidth]{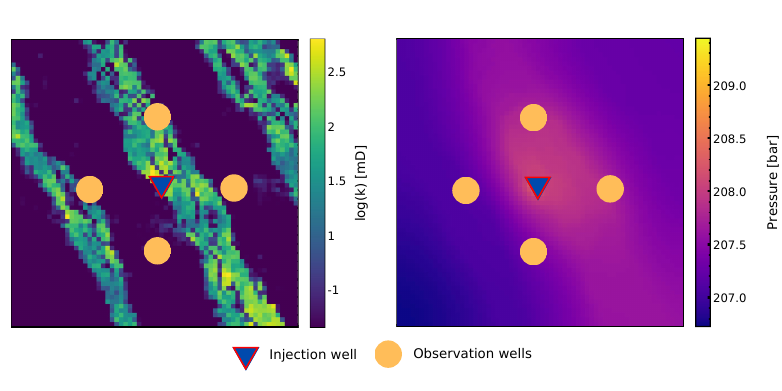}
\caption{Channelized reservoir model (left) and DARTS-simulated pressure field (right) with injection well (triangle) and monitoring wells (circles). The pressure distribution correlates with the high-permeability channels and the injector–producer configuration: it is strongest along the dominant connected channel, with weaker propagation along the two minor channels.}
\label{fig:darts_simulation}
\end{figure}

Figure~\ref{fig:darts_simulation} illustrates the channelized reservoir model and the corresponding pressure field simulated with DARTS. The left panel shows the log-permeability field with high-permeability channels, while the right panel displays the pressure distribution after injection. The injection well (triangle) and four monitoring wells (circles) are indicated. The pressure field clearly follows the channel structure, with higher pressures propagating preferentially along the high-permeability pathways.

\subsection{Diffusion Model Implementation}

The score-based diffusion model is implemented to learn the distribution of channelized permeability fields and generate new realizations that maintain geological realism.

\subsubsection{Neural Network Architecture}

\begin{figure}[p]
\centering
\includegraphics[width=\textwidth]{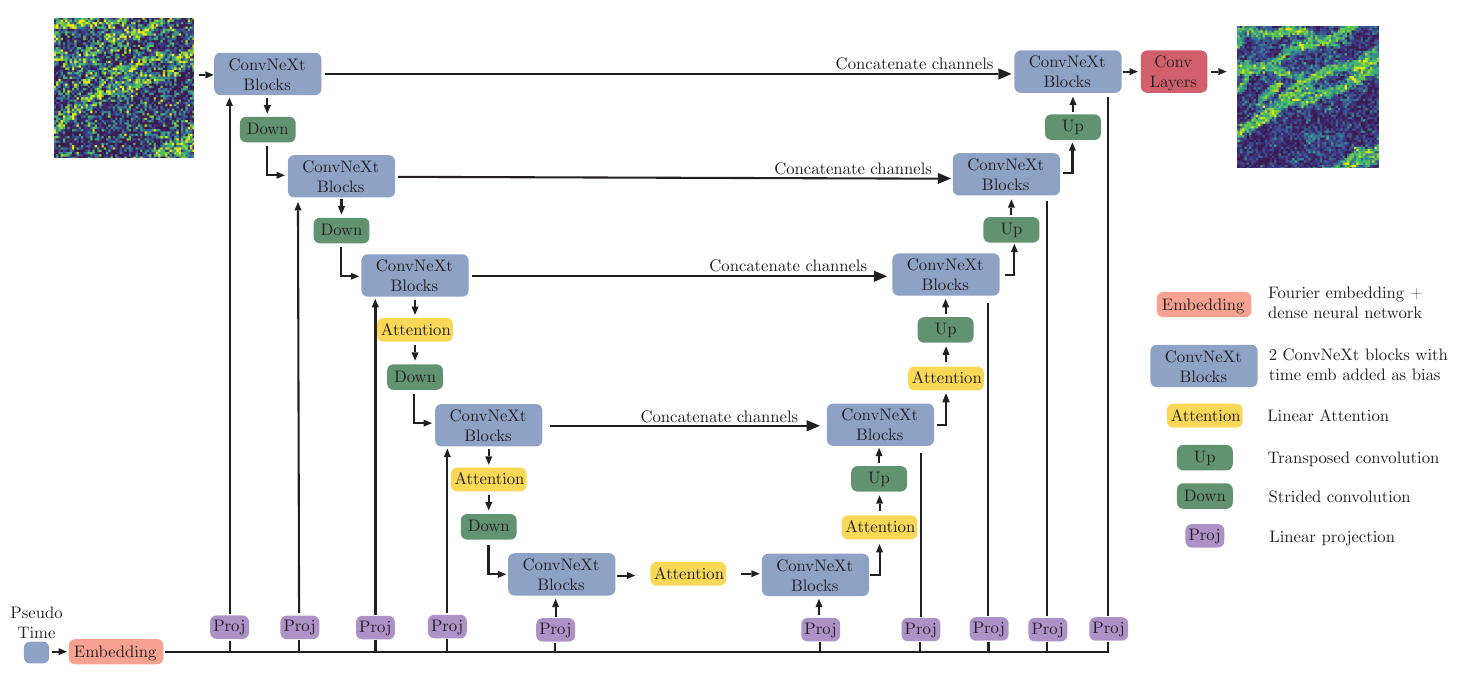}
\caption{Visualization of UNet architecture with attention.}
\label{fig:unet_architecture}
\end{figure}

We employ a hybrid UNet-Transformer \cite{Li2023} architecture that combines convolutional neural networks with efficient attention mechanisms for learning the score function of channelized permeability distributions. The architecture is detailed in Figure \ref{fig:unet_architecture}. We use a U-Net with attention blocks to capture multi-scale features in permeability fields. The network begins with a time embedding module that uses Gaussian Fourier projections to encode the continuous diffusion time into high-dimensional representations. This temporal signal allows the network to adapt its predictions to the current noise level during the diffusion process.

The encoder path consists of four blocks that progressively downsample the input while expanding the feature channels. Each block uses \texttt{ConvNeXt}-style residual blocks \citep{liu2022convnet}: a $7{\times}7$ convolution to capture broader spatial context, followed by $3{\times}3$ convolutions and a residual connection; depthwise separable convolutions are not used unless explicitly stated. The time embeddings are injected at each resolution level, enabling the network to modulate its feature extraction based on the diffusion timestep.

At the bottleneck and intermediate resolutions, we incorporate Spatial Linformer attention modules \citep{wang2020linformer}. Unlike standard self-attention which scales quadratically with input size, Linformer achieves linear complexity through low-rank approximations. This builds on the transformer architecture's success in capturing long-range dependencies \citep{vaswani2017attention} while addressing computational constraints. This is valuable for geological applications where maintaining long-range spatial dependencies—such as channel connectivity across the domain—is important for realistic permeability field generation.

The decoder path mirrors the encoder with skip connections that preserve fine-scale details during upsampling. Through transposed convolutions and ConvNextBlock refinements, the network progressively reconstructs the score function at the original resolution. The final output represents $s_\theta(\mathbf{x}_t, t)$, the estimated score function that guides the reverse diffusion process.

\subsubsection{Training Dataset and Procedure}

The training dataset comprises 3,242 channelized permeability realizations generated using sequential indicator simulation with systematically varied geostatistical parameters. This ensures a comprehensive coverage of plausible geological patterns, with each 64×64 field representing a different realization of channelized formations with varying orientations, widths, and connectivity patterns.

The model training employs standard practices for deep generative models. We use the Adam optimizer with a cosine annealing learning rate schedule that gradually reduces from $10^{-4}$ to $10^{-6}$, promoting stable convergence. The training process minimizes the denoising score matching objective shown in Algorithm \ref{alg:dsm_training}, which teaches the network to predict the gradient of the log-density at various noise levels.

The model converged in 525 epochs. The final loss of 308.76 represents an 83.6\% reduction from the initial value of 1,882.96. The quick convergence shows that channelized patterns are well-structured and the UNet-Transformer architecture captures these geological features effectively.

\begin{algorithm}[h]
\caption{Training Score-Based Diffusion Models}
\label{alg:dsm_training}
\KwIn{Dataset $\mathcal{D} = \{\mathbf{x}_i\}_{i=1}^N$, neural network $s_\theta$, noise schedule $\sigma(t)$}
\KwOut{Trained score network $s_\theta$}
\For{epoch = 1 to $N_{epochs}$}{
  \For{batch $\mathcal{B} \subset \mathcal{D}$}{
    Sample $t \sim \mathcal{U}(0, T)$ uniformly\;
    Sample $\boldsymbol{\epsilon} \sim \mathcal{N}(0, \mathbf{I})$\;
    Compute $\mathbf{x}_t = \mathbf{x}_0 + \sigma_t \boldsymbol{\epsilon}$\;
    Compute loss: $\mathcal{L} = \|\boldsymbol{\epsilon} + \sigma_t s_\theta(\mathbf{x}_t, t)\|^2$\;
    Update $\theta$ using gradient descent on $\mathcal{L}$\;
  }
}
\Return trained $s_\theta$\;
\end{algorithm}

\subsubsection{Sampling Strategy}

The trained diffusion model supports two complementary sampling approaches for generating permeability fields. The Predictor-Corrector (PC) sampler combines Euler-Maruyama steps for the reverse SDE with Langevin MCMC corrections, producing high-quality samples with stochastic diversity. This approach excels in exploring the full distribution of channelized patterns learned during training.

For our ESMDA application, however, we primarily employ the probability flow ODE formulation:
\begin{equation}
\frac{d\mathbf{x}}{dt} = -\frac{1}{2}g^2(t)s_\theta(\mathbf{x}_t, t)
\end{equation}

Deterministic sampling helps, because we can reproduce results exactly for ensemble-based data assimilation. Most importantly, it ensures reproducibility, which is essential for scientific validation and debugging. Given the same initial noise vector, the ODE always produces identical realizations. This property is used for debugging, validation, and ensuring consistent results across different computational environments. The ODE solver employs the midpoint method with 1000 integration steps, balancing numerical accuracy with computational efficiency.


Generated samples are not modified beyond reversing the preprocessing used for training. Specifically, inputs were standardized (per-pixel z-scoring of $\log k$) to ease optimization; at generation time we denormalize the network outputs using the training statistics and invert the log transform to recover permeability in millidarcies. We then compute and report simple diagnostics—channel proportion ($0.40\pm0.05$) and value-range checks ($k\in[10,10^4]$ mD)—but we do not filter or discard samples based on these checks.

\begin{figure}[htbp]
\centering
\includegraphics[width=0.9\textwidth]{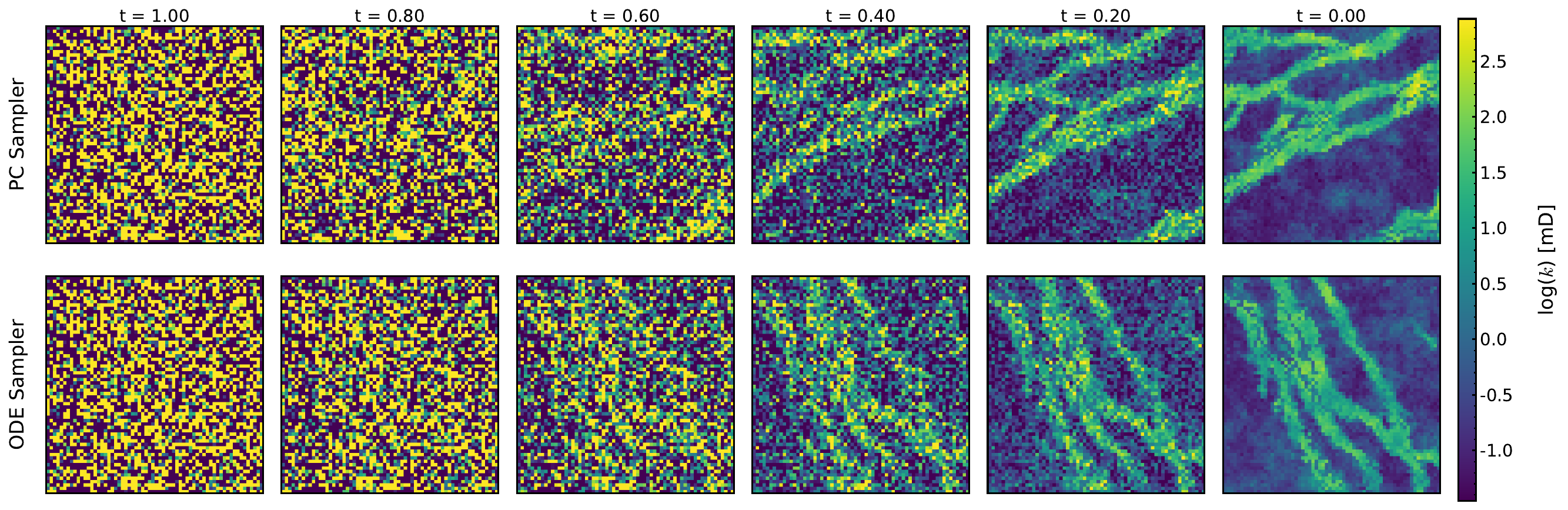}
\caption{Evolution of channelized structures during generation. ODE (top) and PC (bottom) samplers progressively denoise from Gaussian to realistic permeability fields.}
\label{fig:comparative_evolution}
\end{figure}

Figure~\ref{fig:comparative_evolution} illustrates the generation process, showing how both sampling methods progressively transform Gaussian noise into realistic channelized permeability fields while preserving distinct geological features.

\subsection{ESMDA Configuration}

The data assimilation employs ESMDA with four iterations ($N_a = 4$), using uniform inflation coefficients of $\alpha_i = 4$ for all iterations. This choice balances computational efficiency with adequate linearization of the nonlinear inverse problem. The observation error is set to 1\% of the true values, consistent with typical pressure measurement uncertainties. To maintain physical plausibility, parameter bounds constrain the log-permeability values to the range $\log_{10}(k) \in [1, 4]$, corresponding to permeability values between 10 and 10,000 mD. We evaluate the framework performance across multiple ensemble sizes (50, 100, 200, 500, and 1000 members) to understand scaling behavior and practical requirements for operational deployment.

With the inflation schedule $\{\alpha_i\}$ and observation–error covariance $\mathbf{C}_D$ defined above, the localized ESMDA analysis step at iteration $i$ reads:
\begin{equation} \label{eq:esmda_localized}
\mathbf{z}_j^{a} = \mathbf{z}_j^{f} + \mathbf{L} \circ \left[ \mathbf{C}_{zD}^{f} \left( \mathbf{C}_{DD}^{f} + \alpha_i \mathbf{C}_D \right)^{-1} \left( \mathbf{d}_j - \mathbf{G}(\mathbf{z}_j^{f}) \right) \right]
\end{equation}
where $\circ$ denotes the Schur (element-wise) product, $\mathbf{L}$ is the localization matrix, and $\mathbf{d}_j=\mathbf{d}_{\text{obs}}+\sqrt{\alpha_i}\,\boldsymbol{\epsilon}_j$ with $\boldsymbol{\epsilon}_j\sim\mathcal{N}(0,\mathbf{C}_D)$.

\subsection{Machine Learning-Enhanced Localization}

Our ML localization works by training a fast proxy model on the working ensemble to improve covariance estimation. Adapting the machine learning-based distance-free localization introduced in recent studies \citep{silva2025mitigating,silva2025machine}, we tailor these methods for channelized reservoir characterization. Given ensemble pairs $\{(\mathbf{z}_j, \mathbf{d}_j)\}_{j=1}^{N_e}$ from DARTS simulations, we train:

\begin{equation}
f^{ML}: \mathbf{z} \rightarrow \mathbf{d}
\end{equation}

where $\mathbf{z} \in \mathbb{R}^{4096}$ is the flattened log-permeability field and $\mathbf{d} \in \mathbb{R}^{96}$ contains pressure observations.

\subsubsection{Enhanced Covariance Estimation}

The enhanced covariance estimation uses a super-ensemble of $N_s = 5000$ realizations:
\begin{equation}\label{eq:ml_covariance}
\tilde{\mathbf{C}}_{zD}^{ML} = \frac{1}{N_s-1}(\mathbf{Z}_s - \bar{\mathbf{Z}}_s)(\mathbf{D}_s^{ML} - \bar{\mathbf{D}}_s^{ML})^T
\end{equation}
where $\mathbf{D}_s^{ML} = [f^{ML}(\mathbf{z}_1^s), ..., f^{ML}(\mathbf{z}_{N_s}^s)]$ contains ML predictions.

For the ML-enhanced localization, we compute localization coefficients using a modified pseudo-optimal approach. When model parameters lack direct spatial relationships with observations, conventional distance-based schemes become inadequate. We therefore employ an adaptation of the pseudo-optimal taper function introduced by \cite{furrer2007estimation}, selected for its rigorous theoretical foundation and computational efficiency.


Following \citet{furrer2007estimation}, we employ a pseudo-optimal Schur taper that minimizes the expected Frobenius-norm (matrix Euclidean norm) error between the true and the localized covariance. Their weight is changed to the parameter–data cross-covariance \((\mathbf{z},\mathbf{d})\) and replace unknown population quantities with ML-enhanced plug-in estimates computed from the super-ensemble. For entry \((i,j)\) of the localization matrix we set
\begin{equation}\label{eq:ml_localization_fb}
\mathbf{L}_{ij}^{\text{ML}}
= \frac{\big(\tilde{c}_{ij}^{\text{ML}}\big)^2}
       {\big(\tilde{c}_{ij}^{\text{ML}}\big)^2
        + \frac{\big(\tilde{c}_{ij}^{\text{ML}}\big)^2
        + \tilde{c}_{ii}^{\text{ML}}\tilde{c}_{jj}^{\text{ML}}}{N_e}},
\end{equation}
where \(\tilde{c}_{ij}^{\text{ML}}\) is the ML-enhanced covariance between parameter \(i\) and datum \(j\), and \(N_e\) is the ensemble size used in the ESMDA update. To avoid retaining near-zero, noise-dominated terms, we apply a practical thresholding rule: set \(\mathbf{L}_{ij}^{\text{ML}}=0\) whenever \(|\tilde{c}_{ij}^{\text{ML}}| < \eta\sqrt{\tilde{c}_{ii}^{\text{ML}}\tilde{c}_{jj}^{\text{ML}}}\) with \(\eta=10^{-3}\). This plug-in adaptation yields an effective suppression of spurious correlations in high-dimensional, small-ensemble settings.

\begin{algorithm}[h]
\caption{ML-Enhanced Localization}
\label{alg:ml_localization}
\KwIn{Ensemble $\{\mathbf{z}_j, \mathbf{d}_j\}_{j=1}^{N_e}$, ML model type, super-ensemble size $N_s$}
\KwOut{Localization matrix $\mathbf{L}^{ML}$}
\tcp{Step 1: Train ML proxy}
Train $f^{ML}$ using $\{(\mathbf{z}_j, \mathbf{d}_j)\}_{j=1}^{N_e}$ with 80/20 train-validation split\;
\tcp{Step 2: Generate super-ensemble}
\For{$k = 1$ to $N_s$}{
  Sample $\mathbf{z}_k^s$ from prior (diffusion model)\;
  Predict $\mathbf{d}_k^{ML} = f^{ML}(\mathbf{z}_k^s)$\;
}
\tcp{Step 3: Compute enhanced covariance}
Calculate $\tilde{\mathbf{C}}_{zD}^{ML}$ using Equation \ref{eq:ml_covariance}\;
\tcp{Step 4: Calculate localization}
\For{each parameter $i$ and observation $j$}{
  Compute $\mathbf{L}_{ij}^{ML}$ using Equation \ref{eq:ml_localization_fb}\;
}
\Return $\mathbf{L}^{ML}$\;
\end{algorithm}

\subsubsection{Machine Learning Models for State Estimation}

We capture the relationship between permeability fields and pressure responses in reservoir simulation in a tabular dataset, where each grid cell's permeability serves as a feature and well pressures constitute the targets. This structure, combined with recent evidence that classical machine learning methods can outperform deep learning on tabular data \citep{shwartz2022tabular,borisov2022deep,grinsztajn2022tree}, guides our model selection.

We use three different ML models to approximate the nonlinear forward model. Linear regression provides a computationally efficient baseline, establishing whether nonlinearity is essential for accurate predictions. Despite its simplicity, linear models struggle with the complex multiphase flow physics inherent in CO$_2$ injection scenarios.

Random forest models \citep{breiman2001random} address these limitations through ensemble decision trees trained on bootstrap samples. The method naturally handles the sparse influence patterns characteristic of reservoir flow, where distant grid cells often have negligible effect on local pressure observations. By averaging predictions across multiple trees and using random feature selection at each split, random forests capture nonlinear relationships while maintaining robustness against overfitting. XGBoost \citep{chen2016xgboost} extends the ensemble approach through gradient boosting, where each successive tree corrects errors from previous iterations.  

The choice of tree-based methods for our application comes from two data characteristics identified by \citet{grinsztajn2022tree}. First, many grid cells represent uninformative features for specific pressure observations due to flow barriers and distance effects. Second, pressure responses can exhibit non-smooth behavior with sharp transitions at permeability contrasts and channel boundaries. 

These properties explain the performance gap observed in our experiments, where tree-based methods achieve prediction errors 10-15 times lower than linear regression while preserving geological patterns in the localization coefficients. All models employ default hyperparameters from their respective implementations (scikit-learn \citep{pedregosa2011scikit} and XGBoost library \citep{chen2016xgboost}), reflecting operational constraints and ensuring reproducibility.

\subsection{Integrated Workflow}

The workflow consists of two phases: (1) initial ensemble generation using geostatistical models, and (2) iterative ESMDA with ML-enhanced localization. Algorithm \ref{alg:integrated_workflow} details the entire process.

\begin{algorithm}[H]
\caption{Integrated Diffusion-ML-ESMDA Workflow}
\label{alg:integrated_workflow}
\KwIn{Ensemble size $N_e$, observations $\mathbf{d}_\text{obs}$, trained diffusion model, ML model type}
\KwOut{Posterior ensemble $\{\mathbf{z}_j^a\}_{j=1}^{N_e}$}
\tcp{Phase 1: Initial Ensemble Generation}
Generate initial ensemble $\{\mathbf{z}_j^0\}_{j=1}^{N_e}$ using geostatistical models (e.g., FLUVSIM)\;
\tcp{Phase 2: ESMDA with ML-Enhanced Localization}
\For{$i = 1$ to $N_a$ iterations}{
  \tcp{Forward simulations}
  \For{$j = 1$ to $N_e$}{
    Run DARTS: $\mathbf{d}_j = \mathbf{G}(\mathbf{z}_j^{i-1})$\;
  }
  \tcp{ML proxy training}
  Train ML model: $f^{ML} = \text{fit}(\{\mathbf{z}_j^{i-1}\}, \{\mathbf{d}_j\})$\;
  \tcp{Super-ensemble generation}
  Generate $\{\mathbf{z}_k\}_{k=1}^{N_s}$ using diffusion model\;
  Predict: $\mathbf{d}_k^{ML} = f^{ML}(\mathbf{z}_k)$ for $k = 1, \ldots, N_s$\;
  \tcp{Enhanced localization}
  Compute $\tilde{\mathbf{C}}_{zD}^{ML}$ using super-ensemble\;
  Calculate localization matrix $\mathbf{L}^{ML}$ (Algorithm \ref{alg:ml_localization})\;
  \tcp{ESMDA update}
  \For{$j = 1$ to $N_e$}{
    Perturb observations: $\mathbf{d}_j^{\text{pert}} = \mathbf{d}_\text{obs} + \sqrt{\alpha_i}\boldsymbol{\epsilon}_j$\;
    Apply localized update to get $\mathbf{z}_j^i$\;
  }
}
\Return Posterior ensemble $\{\mathbf{z}_j^{N_a}\}_{j=1}^{N_e}$\;
\end{algorithm}

The framework's modular design allows flexibility in component selection. Different ML models can be employed based on problem characteristics—Random Forest for capturing local effects, XGBoost for handling complex nonlinearities, or Linear Regression when computational resources are extremely limited. Similarly, the diffusion model architecture can be adapted to different geological environments by training on appropriate datasets, be they from channelized, fractured, or layered systems.


\begin{figure}[htbp]
\centering
\includegraphics[width=0.95\textwidth]{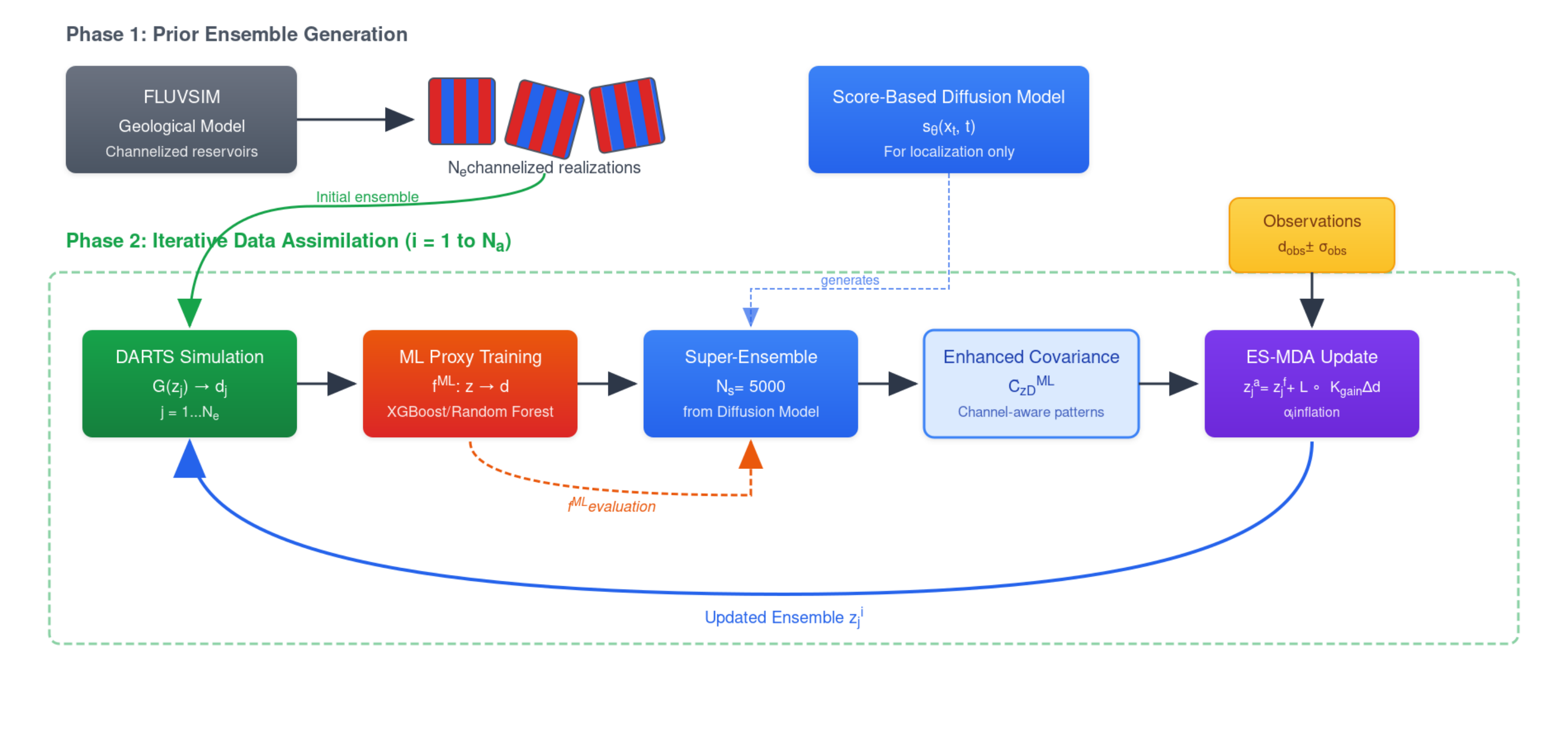}
\caption{Integrated workflow combining score-based diffusion models, ML-based localization, and ESMDA. Phase 1: prior ensemble generation via geostatistical models. Phase 2: iterative data assimilation with ML-enhanced localization.}
\label{fig:integrated_workflow}
\end{figure}

\section{Results and Analysis}\label{section:results}
This section presents the evaluation of our integrated framework for uncertainty quantification in geological carbon storage. We assess the diffusion model's performance, analyze machine learning proxy models for localization coefficient prediction, and evaluate the data assimilation results across different ensemble sizes. The focus is on how well the framework captures channelized permeability structures and improves pressure predictions through ML-enhanced localization. To quantify variance preservation, we report the normalized variance (NV), defined as the ratio between the posterior and prior ensemble variances of log–permeability: $\mathrm{NV} = \mathrm{tr}(\mathbf{C}_{\text{post}}) / \mathrm{tr}(\mathbf{C}_{\text{prior}})$, where $\mathbf{C}_{\text{post}}$ and $\mathbf{C}_{\text{prior}}$ are the ensemble covariance matrices over all grid cells. An NV of 1 indicates that there is no variance loss, while smaller values indicate that there is a reduced variance.

In what follows, we evaluate each localization method using three metrics: (i) the normalized variance (NV) of log-permeability—our primary metric, which indicates how well the ensemble avoids collapse; (ii) data-match accuracy measured by the RMSE between simulated and observed pressures; and (iii) spatial diagnostics (localization maps and posterior permeability fields) to assess geological realism of the updates.

\subsection{Generating Geological Models with Diffusion}

The diffusion model converged quickly during training on 3,242 channelized permeability realizations. Starting from an initial loss of 1,882.96, the model achieved stable convergence at 308.76 after only 525 epochs—representing an 83.6\% reduction in loss and early termination compared to the planned 5,000 epochs. The optimal model checkpoint, captured at epoch 515 with a loss of 306.46, was selected for all subsequent experiments.

\begin{figure}[htbp]
\centering
\includegraphics[width=\textwidth]{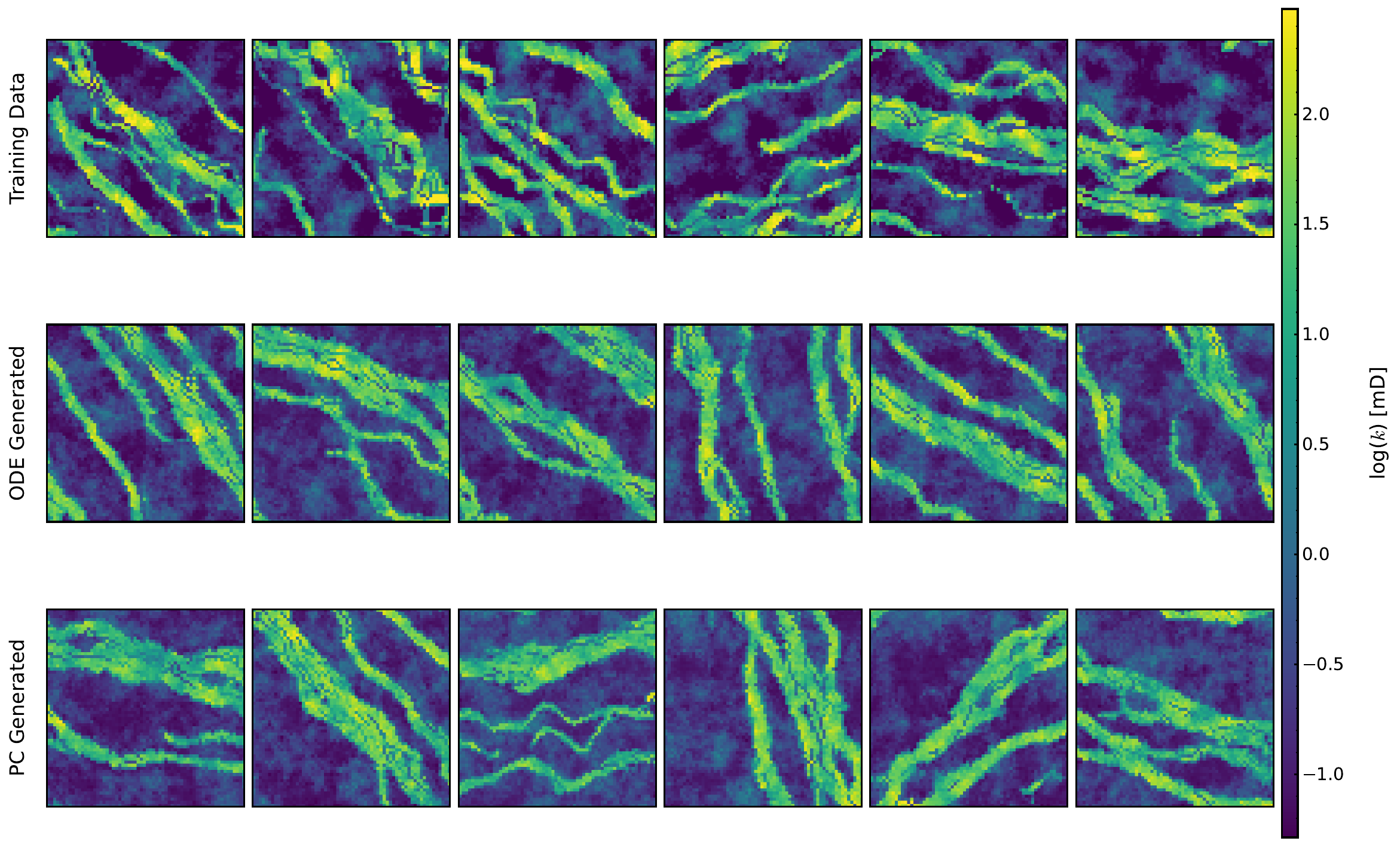}
\caption{Visual comparison of training data (top row) with generated realizations using ODE sampling (middle row) and SDE sampling (bottom row).}
\label{fig:training_vs_generated}
\end{figure}

Figure~\ref{fig:training_vs_generated} demonstrates the quality of generated permeability fields compared to the original FLUVSIM training data. The comparison shows realizations generated using both ODE and SDE sampling approaches alongside true training samples.

Both samplers generate geologically plausible variations without replicating training samples. In practice, they produce diverse permeability realizations that capture channel geometry and connectivity, which we then use to sample geological models for the data-assimilation experiments.

\subsection{Machine Learning Proxy for Localization}

We evaluate three machine learning algorithms for predicting localization coefficients: linear regression, random forest, and XGBoost. Table~\ref{tab:ml_performance} summarizes their performance across different ensemble sizes, revealing distinct patterns that inform our framework design.

\begin{table}[htbp]
\centering
\caption{Machine Learning Model Performance for Localization Coefficient Prediction}
\label{tab:ml_performance}
\begin{subtable}{\textwidth}
\centering
\caption{Root Mean Square Error (RMSE)}
\small
\setlength{\tabcolsep}{5pt}
\begin{tabular}{lccccc}
\toprule
\textbf{Model} & \multicolumn{5}{c}{\textbf{Number of Ensemble Members ($N_e$)}} \\
\cmidrule(lr){2-6}
& 50 & 100 & 200 & 500 & 1000 \\
\midrule
Linear Regression & 2.351 & 2.828 & 0.681 & 1.262 & 0.268 \\
Random Forest & 0.254 & 0.175 & 0.191 & 0.177 & 0.159 \\
XGBoost & 0.310 & 0.198 & 0.216 & 0.188 & 0.159 \\
\bottomrule
\end{tabular}
\end{subtable}

\vspace{0.5em}

\begin{subtable}{\textwidth}
\centering
\caption{Training Time (seconds)}
\small
\setlength{\tabcolsep}{5pt}
\begin{tabular}{lccccc}
\toprule
\textbf{Model} & \multicolumn{5}{c}{\textbf{Number of Ensemble Members ($N_e$)}} \\
\cmidrule(lr){2-6}
& 50 & 100 & 200 & 500 & 1000 \\
\midrule
Linear Regression & 0.016 & 0.027 & 0.067 & 0.141 & 0.40 \\
Random Forest & 3.528 & 6.238 & 15.678 & 38.890 & 106.8 \\
XGBoost & 1.301 & 1.849 & 4.372 & 15.767 & 70.9 \\
\bottomrule
\end{tabular}
\end{subtable}
\end{table}

Tree-based methods (Random Forest and XGBoost) consistently generate better results than linear regression, achieving RMSE values below 0.31 across the full range of ensemble sizes. The much less consistent RMSE values of the linear regression (RMSE values ranging from 0.268 to 2.828) illustrate that this method is not as effective in capturing the pressure response in the strongly heterogeneous channelized systems.

While Random Forest achieves the lowest prediction errors for most ensemble sizes, computational cost analysis reveals important practical considerations. Training time for Random Forest increases with ensemble size, reaching 106.8 seconds for $N_e = 1000$. XGBoost offers an attractive alternative, providing comparable accuracy (RMSE within 0.02 of Random Forest) with 33\% lower computational cost for large ensembles.

\subsection{Data Assimilation Performance}

\begin{figure}[htbp]
\centering
\begin{subfigure}{0.48\textwidth}
    \centering
    \includegraphics[width=\textwidth]{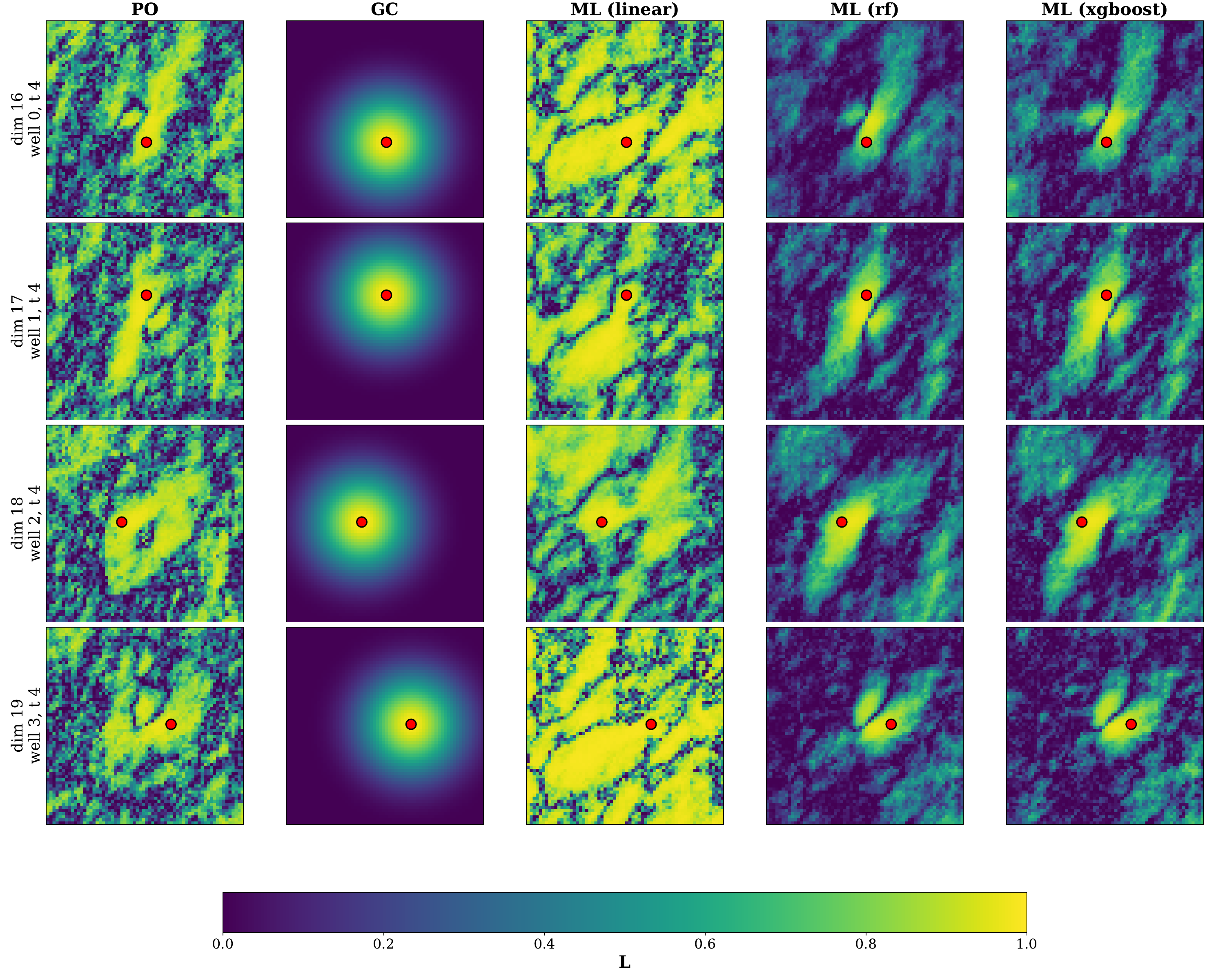}
    \caption{$N_e = 200$}
\end{subfigure}
\hfill
\begin{subfigure}{0.48\textwidth}
    \centering
    \includegraphics[width=\textwidth]{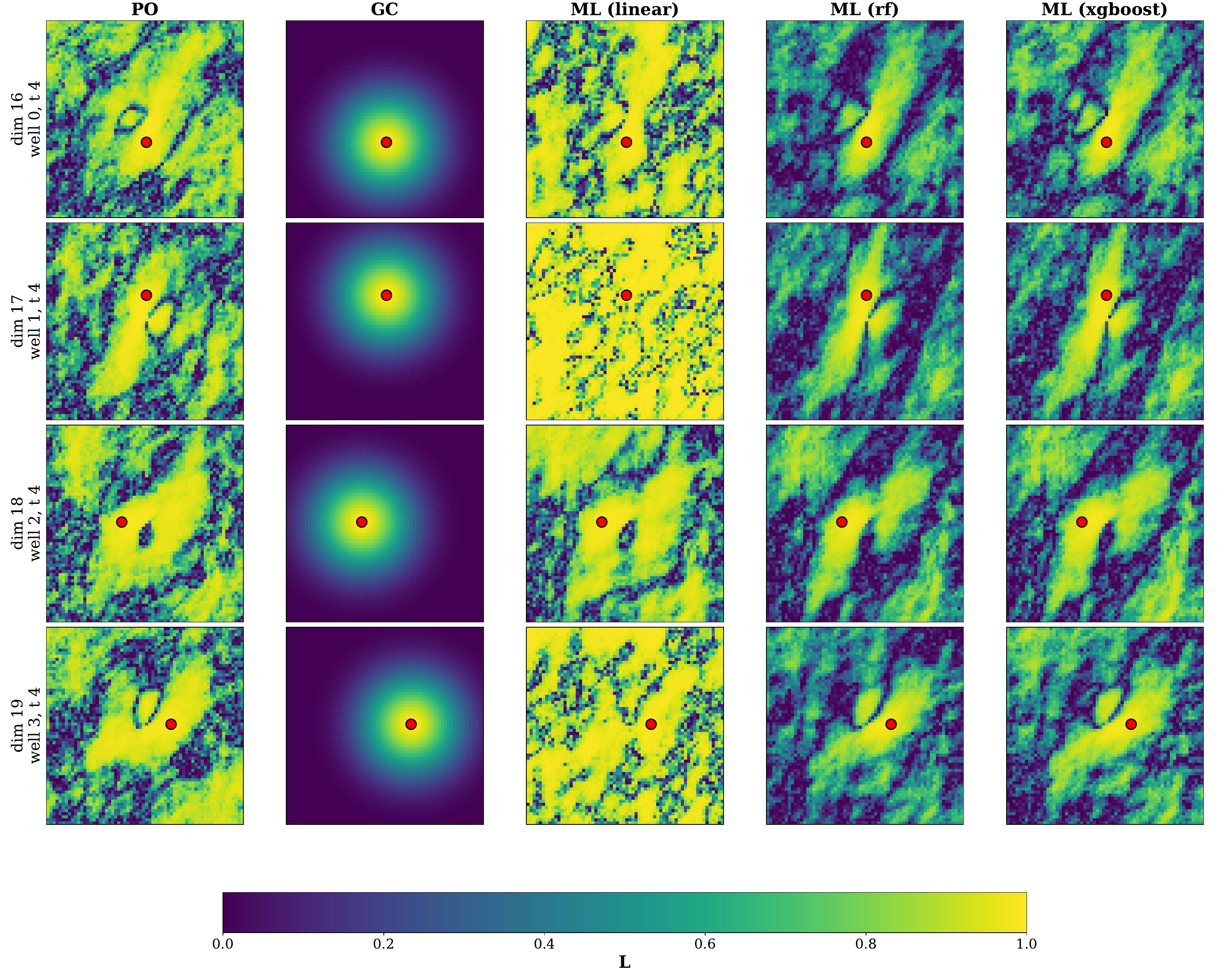}
    \caption{$N_e = 500$}
\end{subfigure}
\caption{Representative localization patterns for medium and large ensemble sizes showing channel-aligned parameter updates for different methods. The data point location is marked with a red circle. Unlike traditional distance-based methods (GC) that produce circular patterns, ML-enhanced localization (ML methods) adapts to the channelized permeability structure, concentrating updates along high-permeability pathways.}
\label{fig:localization_examples}
\end{figure}

ML localization's main benefit shows up when examining the spatial patterns of parameter updates. Figure~\ref{fig:localization_examples} illustrates representative localization patterns for medium ($N_e = 200$) and large ($N_e = 500$) ensemble sizes, demonstrating how ML-enhanced methods adapt to the channelized reservoir structure.

\begin{figure}[htbp]
\centering
\includegraphics[width=\textwidth]{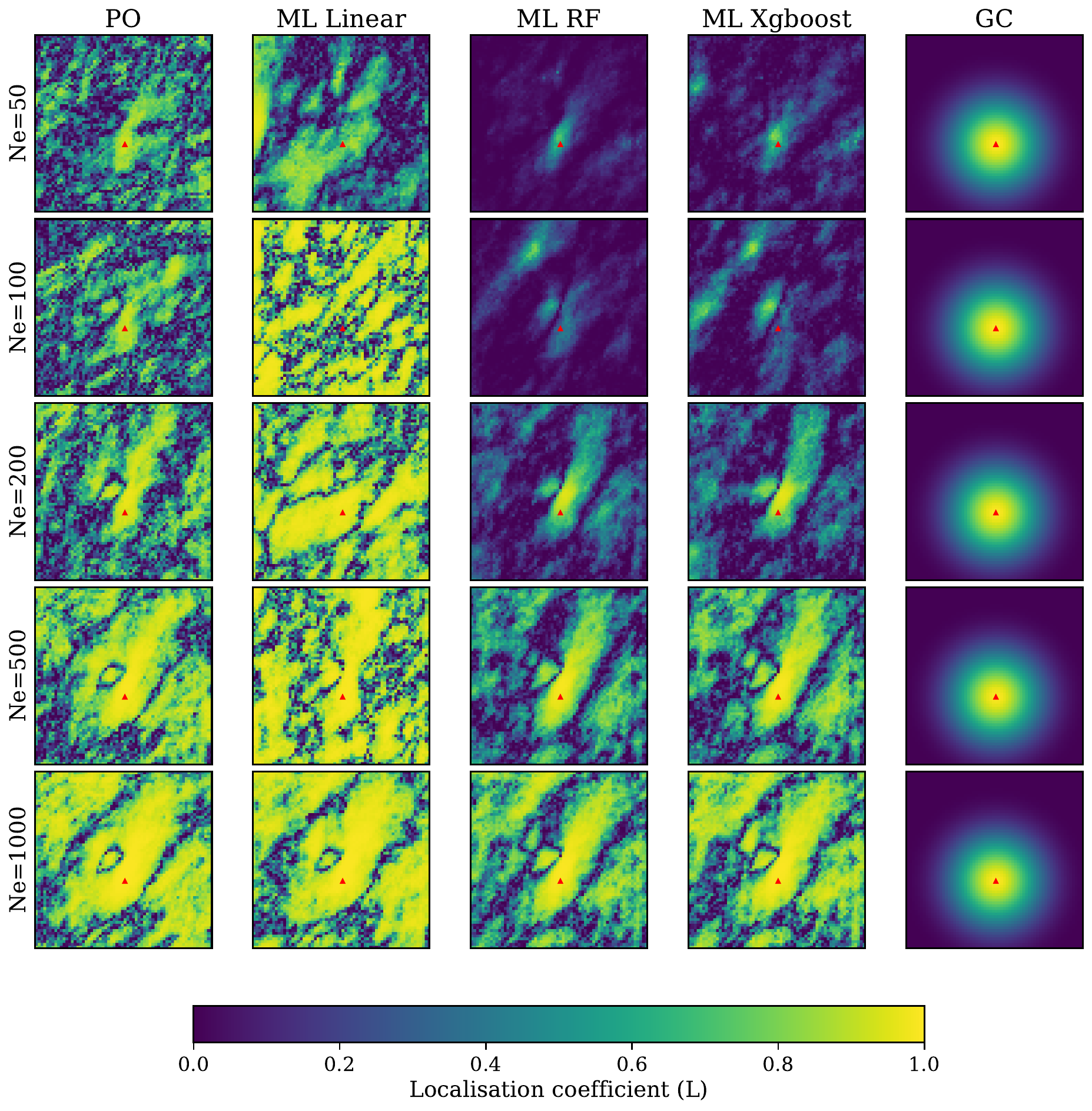}
\caption{Localization patterns for different ensemble sizes ($N_e$ = 50-1000) and methods. ML methods produce channel-aligned patterns; traditional methods show circular patterns.}
\label{fig:localization_maps_compare}
\end{figure}

A comparison across all ensemble sizes and methods is presented in Figure~\ref{fig:localization_maps_compare}.

Traditional distance-based localization (Gaspari-Cohn) produces circular update regions centered on observation wells, indiscriminately modifying both channel and background permeabilities. This approach fails to respect the fundamental geological structure where pressure information propagates preferentially along high-permeability channels. In contrast, ML-enhanced methods generate updates that follow channel geometries, concentrating modifications where they are geologically meaningful. This channel-aligned updating is maintained consistently across all ensemble sizes, demonstrating the robustness of the ML approach.

Further details and additional localization maps for the remaining ensemble sizes ($N_e = 50, 100, 1000$) are provided in ~\ref{appendix:localization_maps}, which illustrates the spatial features of the localization coefficients for each case.

\begin{figure}[htbp]
\centering
\includegraphics[width=\textwidth]{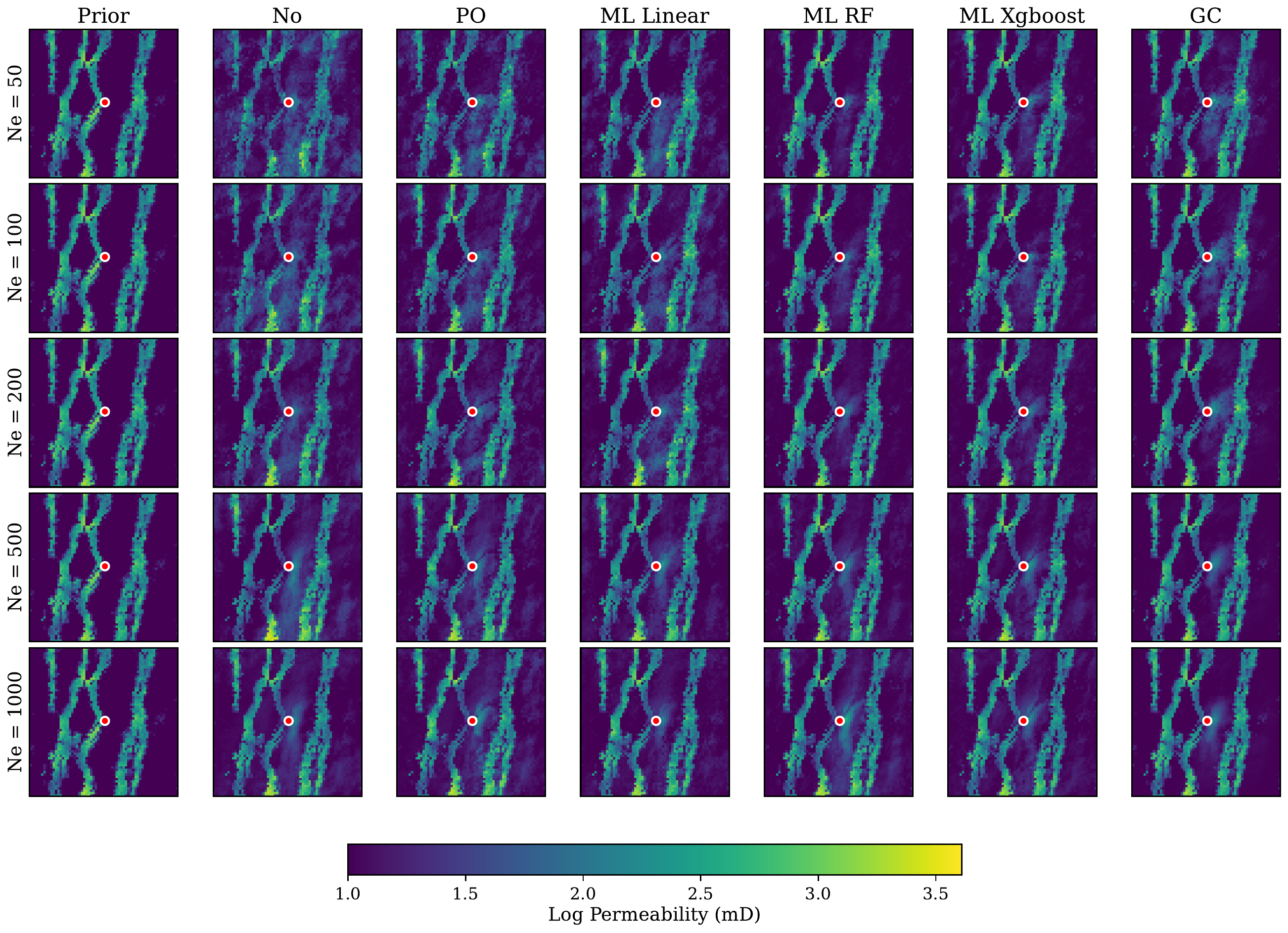}
\caption{Sample of posterior permeability fields for different ensemble sizes ($N_e$) and localization methods. ML-based methods preserve channel structures better than other methods, especially for small ensembles.}
\label{fig:posterior_perms_compare}
\end{figure}

The posterior permeability fields for the different localization strategies are illustrated in Figure~\ref{fig:posterior_perms_compare}. The figure displays a representative posterior permeability sample for different ensemble sizes and localization methods. For small ensemble sizes ($N_e=50$), the absence of localization or the use of pseudo-optimal localization leads to severely blurred permeability fields. The updates affect the background facies indiscriminately, and even the channel structures lose their sharpness. In contrast, the ML-based localization methods (Random Forest and XGBoost) preserve the channel connectivity and overall geological structure remarkably well, even with a small ensemble. The updates exhibit a channel-like form, which is a direct consequence of the geology-aware localization patterns shown on Figure~\ref{fig:localization_examples} and Figure~\ref{fig:localization_maps_compare}. The Gaspari-Cohn localization produces updates that are distinctly circular, reflecting its distance-based nature and leading to less geologically consistent modifications. As the ensemble size increases, the benefits of ML-based localization become less pronounced, as expected. However, even for larger ensembles ($N_e=500, 1000$), the permeability fields generated with ML-localization exhibit the channelized structure more realistically compared to other methods. For large ensembles, ML-enhanced localization still preserves fine-scale geological detail better than distance-based tapers or no localization, although the margin of improvement is smaller than at low $N_e$. 

\begin{figure}[htbp]
  \centering
  \includegraphics[width=\textwidth]{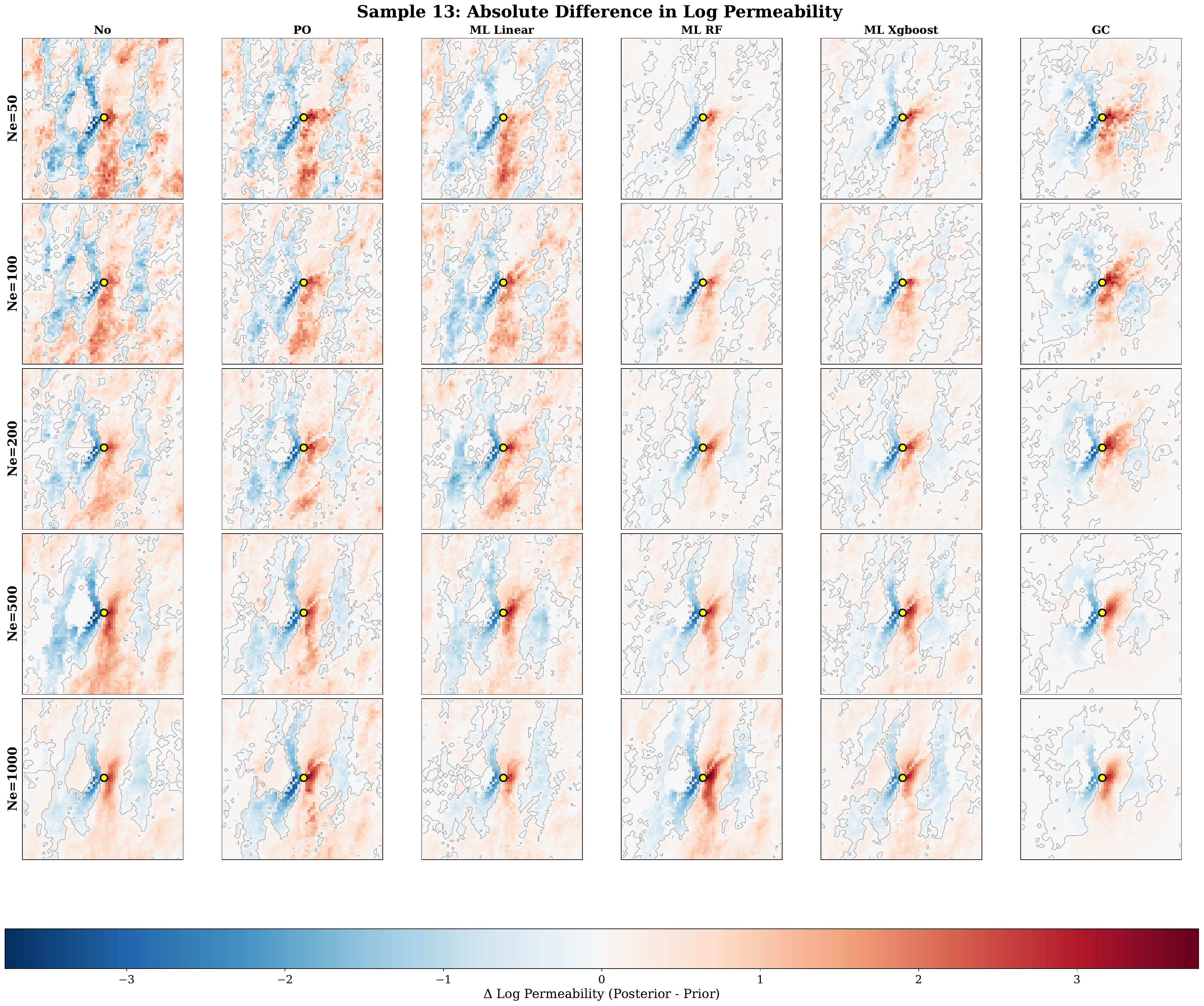}
  \caption{Log-permeability change (posterior – prior) as a function of ensemble size $N_e$ and localization method for the following methods: PO, linear machine learning (ML-linear), random forest (ML RF), xgboost (ML XGBoost)}
  \label{fig:delta_logk_sample13}
\end{figure}

To visualize \emph{where} and \emph{how strongly} each method actually modifies the model, Fig.~\ref{fig:delta_logk_sample13}
shows the absolute change in log–permeability (posterior – prior) for the same realization showed in Fig.~\ref{fig:posterior_perms_compare}.  
The machine-learning localizations concentrate updates along the high-permeability channel that connects to the
observation well (green marker), whereas the un-localised cases disperse changes into the
background. This effect is larger for $N_e$=50 and 100 members. In these cases, the Random Forest– and XGBoost–based localizations concentrate updates along connected channels and preserve ensemble variance better than pseudo-optimal, Gaspari–Cohn, or no localization; the linear ML variant performs in between the tree-based and classical methods.

\begin{figure}[htbp]
\centering
\includegraphics[width=\textwidth]{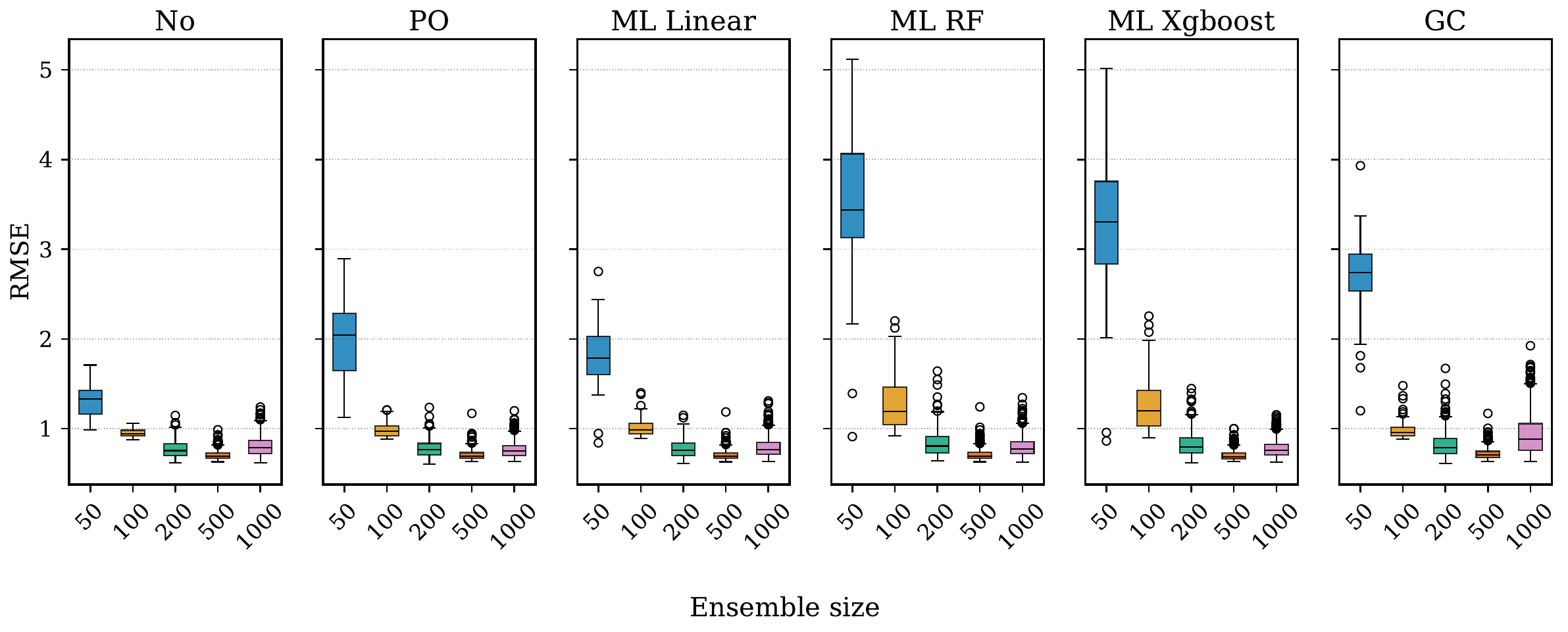}
\caption{RMSE comparison across ensemble sizes.}
\label{fig:rmse_compare}
\end{figure}

To quantify the quality of data matching across different ensemble sizes, Figure~\ref{fig:rmse_compare} presents the RMSE values achieved by each method, providing a comprehensive measure of data matching quality.

The RMSE analysis reveals distinct performance patterns across ensemble sizes. For larger ensembles ($N_e \geq 100$), all localization methods achieve acceptable data matching quality. However, very small ensembles ($N_e = 50$) show degraded performance across all methods, suggesting this ensemble size may be insufficient for reliable data assimilation in channelized systems, even with enhanced localization techniques. Despite these limitations at $N_e = 50$, ML-enhanced methods consistently outperform traditional approaches within each ensemble size category.  

\begin{figure}[htbp]
\centering
\includegraphics[width=\textwidth]{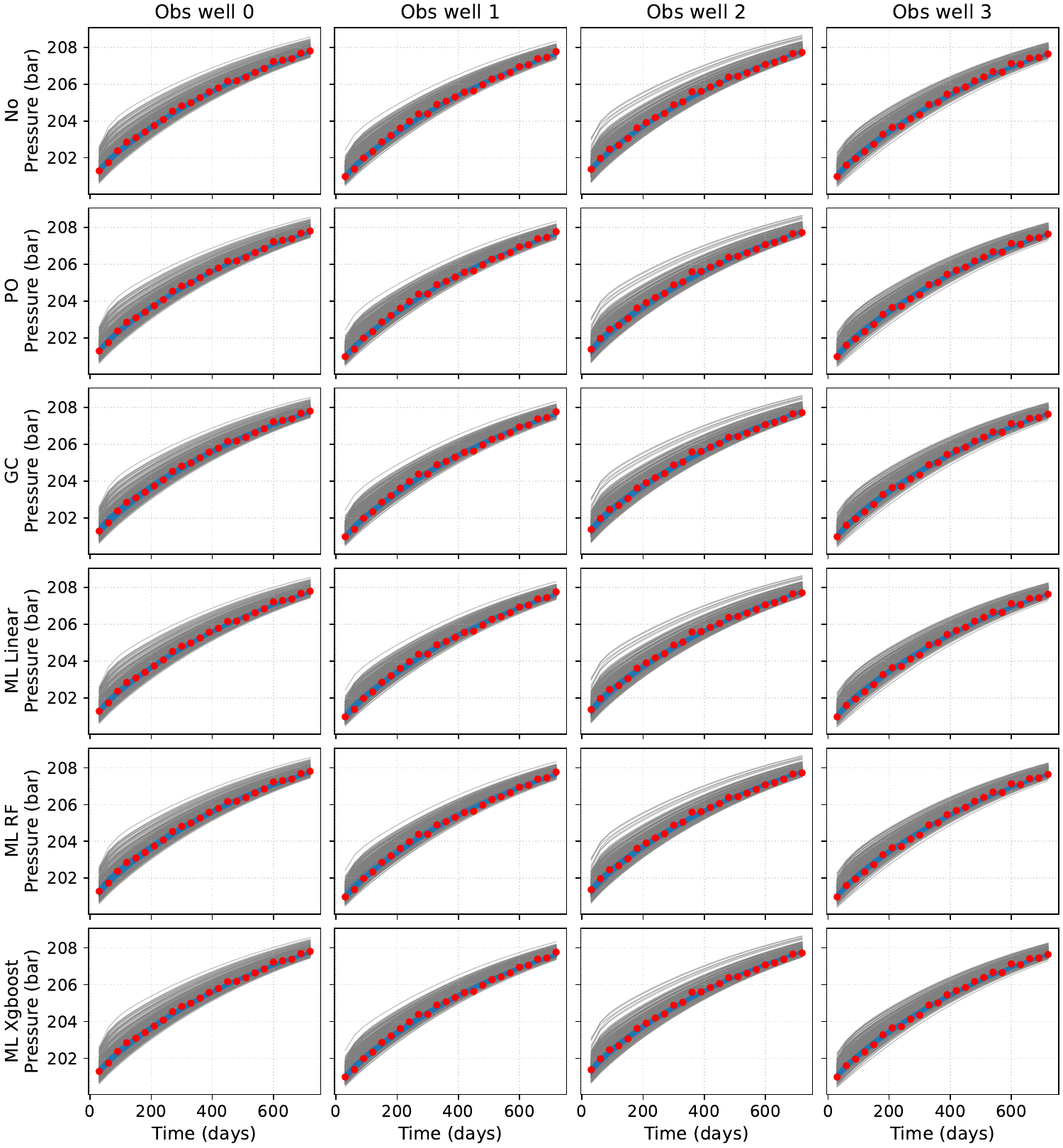}
\caption{Data match for $N_e = 500$ across four monitoring wells. True data (red), posterior (blue), prior (gray) curves.}
\label{fig:data_match_ne500}
\end{figure}

To illustrate the history matching in detail, Figure~\ref{fig:data_match_ne500} presents pressure evolution at four monitoring wells throughout the 10-year CO$_2$ injection period for the $N_e = 500$ case, comparing different localization methods. All methods track the observed pressure evolution, with ensemble mean predictions closely following the true data.

\begin{figure}[htbp]
\centering
\includegraphics[width=\textwidth]{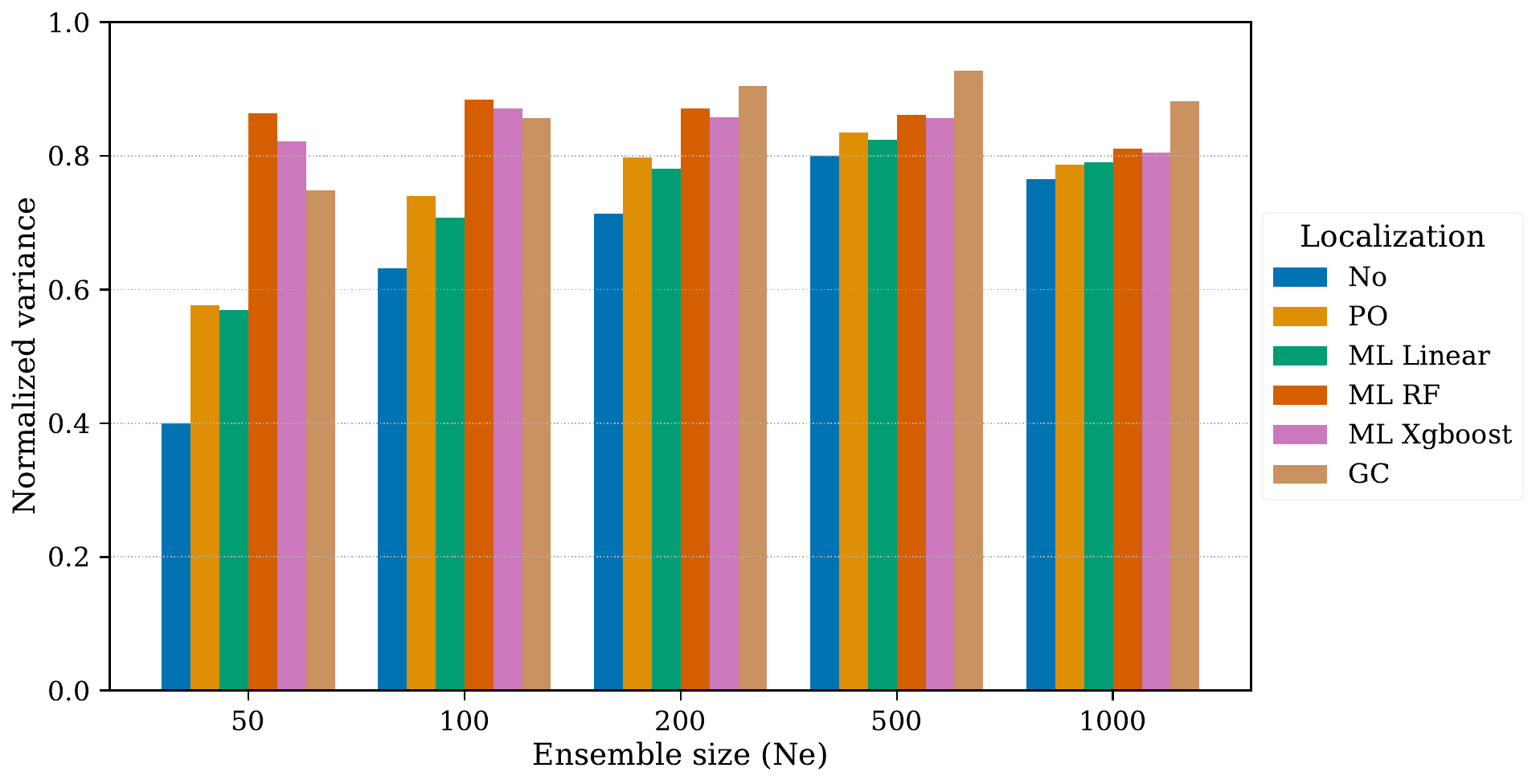}
\caption{Normalized variance comparison across ensemble size for each of the different localization approaches.}
\label{fig:nv_compare}
\end{figure}

The effectiveness of different localization methods in preventing ensemble collapse is quantified through normalized variance (NV), our primary performance metric. The higher the NV, the better the variance is preserved in the data assimilation and the less likely the occurrence of ensemble collapse. Table~\ref{tab:comprehensive_results} and Figure~\ref{fig:nv_compare} present comprehensive results across all tested configurations.

\begin{table*}[htbp]
\centering
\caption{Normalized Variance (NV) Results for Different Localization Methods Across Ensemble Sizes}
\label{tab:comprehensive_results}
\begin{tabular}{lccccc}
\toprule
Method & $N_e$=50 & $N_e$=100 & $N_e$=200 & $N_e$=500 & $N_e$=1000 \\
\midrule
No Localization & 0.400 & 0.632 & 0.714 & 0.801 & 0.765 \\
Pseudo-Optimal & 0.577 & 0.740 & 0.797 & 0.835 & 0.787 \\
Gaspari-Cohn & 0.749 & 0.857 & 0.904 & 0.927 & 0.882 \\
ML-Linear & 0.570 & 0.708 & 0.780 & 0.825 & 0.791 \\
ML-Random Forest & 0.864 & 0.885 & 0.871 & 0.861 & 0.811 \\
ML-XGBoost & 0.821 & 0.872 & 0.858 & 0.857 & 0.805 \\
\bottomrule
\end{tabular}
\end{table*}

Random Forest and XGBoost have the highest NV across all ensemble sizes. The improvements are most pronounced for small ensembles where sampling errors are severe. For $N_e = 50$, ML-Random Forest achieves NV = 0.864 compared to only 0.400 without localization. This is a 116\% improvement that transforms a collapsed ensemble into one that preserves its variance.

The framework exhibits distinct variance preservation across different ensemble size regimes:

\textbf{Small ensembles ($N_e < 100$):} ML-enhanced localization may be necessary to prevent variance collapse. Without proper localization, small ensembles lose the channelized geological structure and underestimate uncertainty. ML methods (RF, ....) restore viable uncertainty quantification with minimal computational overhead.

\textbf{Medium ensembles ($N_e = 100-500$):} ML methods provide benefits, with variance preservation improvements of 20-40\% over no localization. For $N_e = 200$, ML methods achieve NV values of 0.86-0.87 compared to 0.71 without localization, maintaining channelized structures throughout the assimilation process.

\textbf{Large ensembles ($N_e > 500$):} Benefits diminish as natural sampling errors decrease, though ML methods (RF and XGBoost) still show marginal improvements. For $N_e = 1000$, all methods achieve acceptable variance preservation (NV $>$ 0.78), with ML approaches reaching 0.81.

\section{Discussion}\label{section:discussion}

This study investigates whether a combination of score-based diffusion models and machine learning proxies can overcome two persistent drawbacks of ensemble smoothers applied to channelized CO$_2$ storage: systematic variance loss caused by limited ensemble sizes and the geologically unrealistic permeability patterns imposed by distance-based localization. The numerical experiments confirm that the proposed workflow makes progress on both fronts. This represents a step beyond traditional ensemble methods \citep{emerick2013esmda,evensen2019efficient}, which can suffer from variance underestimation and spurious correlations due to limited ensemble sizes. By leveraging machine learning proxy models, such as those recently explored by \citet{lacerda2021machine}, \citet{silva2025mitigating}, and \citet{silva2025machine}, our methodology effectively suppresses spurious correlations in data assimilation.

Across all ensemble sizes, the machine-learning localizations based on Random Forests and XGBoost preserved more prior variance than linear model, pseudo-optimal or Gaspari-Cohn. With just one or two hundred realizations, the additional variance reached 20–40\%, a range that aligns with the ensemble sizes routinely used in practice. The advantage narrowed as the ensemble grew, which is consistent with the expected reduction of spurious correlations at large $N_e$.

One point deserves emphasis: in our workflow, the machine-learning proxies are never used as replacements for the flow simulator inside the Kalman update itself; they serve exclusively to estimate the localization matrix. This design choice matters because localization depends only on the relative structure of the parameter–data covariance, not on the exact reproduction of the pressure traces. Consequently, even proxies that would be inadequate as full surrogate models prove perfectly adequate—and far cheaper—to guide localization. By confining the proxy's role to this task, we avoid the risk of biasing the analysis with approximate physics while still reaping the statistical benefit of a much larger effective ensemble.

Equally important is the qualitative improvement in the spatial structure of the updates. Visual inspection of the localization maps and the posterior permeability fields shows that the machine-learning approach concentrates corrections along high-permeability channels and generates less perturbations on the low-permeability background. Distance-based kernels, by contrast, produce circular footprints that severely underestimate geological connectivity.


The computational overhead of ML-localization is small. Training the proxy at each assimilation step adds a few seconds for ensembles up to 200 members and less than 30 seconds for 500 members, which is minor relative to even a single forward DARTS run, which for this case takes about 2-3 minutes of wall-clock time per realization on one modern CPU core.

The ML-localization method has limitations: We only tested two-dimensional scenarios, leaving the potential computational complexities of three-dimensional geological formations unexplored. The extension to 3D would require addressing memory constraints for diffusion model training. The current GPU architectures limit practical grid sizes to approximately $128^3$ for full-resolution models. Given that most operational reservoir models comprise thousands of grid cells, this limitation needs to be addressed to apply ML-localization in an operational setting. Additionally, our analysis focuses on pressure data integration, neglecting other geophysical data types such as seismic or electromagnetic observations, commonly incorporated in multi-physics integration studies. The reliance on offline training of ML proxy models at each assimilation iteration also poses scalability and flexibility challenges, which asks for further exploration of online learning strategies.

\section{Conclusions}\label{section:conclusions}

In this work, we combine score-based diffusion models, machine-learning proxies, and the Ensemble Smoother with Multiple Data Assimilation (ESMDA) to deliver a variance-preserving, channel-preserving history-matching strategy for channelized CO$_2$ storage reservoirs. The diffusion model generates thousands of geologically plausible permeability fields at negligible cost once trained. The ML proxies exploit these large synthetic ensembles to build more reliable covariance estimates, and the resulting localization mitigates ensemble collapse without additional flow simulations. In tests with two-dimensional channel systems, the approach preserved up to 40\% more variance than standard distance-based tapers, maintained channel connectivity in the posterior fields, and introduced only minor computational overhead.

These results demonstrate that data-driven localization can significantly reduce the gap between small ensembles that are affordable in practice andlarge ensembles required for accurate covariance estimation. Extending the framework to three dimensions, adding multi-physics data streams, and adopting online learning for  the proxies  (i.e., realtime development of ML models as the model runs forward) in a data assimilation framework with ML-localization are the logical next steps toward real-time, uncertainty-aware management of large-scale CO$_2$ storage projects.

\section{Code and data availability}

The source code for all experiments is openly available at
\href{https://github.com/gabrielserrao/DA-with-generative-ml}{github.com/gabrielserrao/DA-with-generative-ml}. The dataset used in this study is archived at
\href{https://doi.org/10.4121/a8ad7808-b923-4335-ba7a-898c8c1232be}{10.4121/a8ad7808-b923-4335-ba7a-898c8c1232be}.

\section*{Acknowledgements}

The authors are grateful to Petróleo Brasileiro S.A.\ (Petrobras) for sponsoring the doctoral research of Gabriel Serr{\~a}o Seabra.

\section*{CRediT authorship contribution statement}

\textbf{Gabriel Serr{\~a}o Seabra}: Conceptualisation, Methodology, Software, Validation,  Data curation, Formal analysis, Investigation, Visualisation, Writing – original draft.  

\textbf{Nikolaj T. M{\"u}cke}: Conceptualisation, Methodology, Software, Validation, Data curation, Formal analysis,  Writing – original draft.

\textbf{Vinicius Luiz Santos Silva}: Conceptualisation, Methodology, Validation, Writing – review \& editing.  

\textbf{Alexandre A. Emerick}: Methodology, Investigation, Writing – review \& editing.  

\textbf{Denis Voskov}: Conceptualisation, Supervision, Writing – review \& editing.  

\textbf{Femke Vossepoel}: Conceptualisation, Methodology, Supervision, Project administration, Funding acquisition, Writing – review \& editing.

\section*{Declaration of competing interest}

The authors declare that they have no known competing financial or personal interests that could have influenced the work reported in this paper.  
Gabriel Serr{\~a}o Seabra, Alexandre Emerick and Vinicius Luiz Santos Silva prepared this manuscript while employed by Petrobras; the company had no role in study design, data collection, data analysis, decision to publish, or manuscript preparation.

\section*{Declaration of generative AI and AI-assisted technologies in the writing process}
During manuscript preparation the authors employed OpenAI's ChatGPT strictly to improve wording and clarity, and Grammarly\textsuperscript{\textregistered} for grammar checking.  After using these tools, the authors reviewed and edited all suggestions to ensure accuracy and originality.  The authors take full responsibility for the content of this publication.


\bibliography{references-checked}


\appendix

\section{Posterior Diffusion Sampling for Hard Data Conditioning}\label{appendix:posterior}

While the main data assimilation workflow presented in this paper uses ESMDA for dynamic data integration, the trained diffusion model also enables direct conditioning on hard permeability data through posterior sampling. This capability is valuable when direct permeability observations are available from well logs, core samples, or other sources. This appendix presents the foundation and validation of the posterior diffusion sampling approach.

\subsection{Theoretical Foundation}

For Bayesian inversion, we need samples from the posterior distribution:

\begin{equation}
p(\mathbf{x} | \mathbf{y}) = \frac{p(\mathbf{y} | \mathbf{x}) p(\mathbf{x})}{p(\mathbf{y})} \propto p(\mathbf{y} | \mathbf{x}) p(\mathbf{x})
\end{equation}

The observations are related to the permeability field through:

\begin{equation}
\mathbf{y} = H(\mathbf{x}) + \boldsymbol{\eta}, \quad \boldsymbol{\eta} \sim \mathcal{N}(0, \sigma_{\text{obs}}^2\mathbf{I})
\end{equation}

For direct permeability measurements, $H$ is a selection matrix extracting values at measurement locations. The posterior score can be decomposed as:

\begin{equation} \label{eq:posterior_score}
\nabla_{\mathbf{x}_t}\log p_t(\mathbf{x}_t | \mathbf{y}) = \underbrace{\nabla_{\mathbf{x}_t}\log p_t(\mathbf{x}_t)}_{\text{Prior score, } s_{\theta}} + \underbrace{\nabla_{\mathbf{x}_t}\log p_t(\mathbf{y} | \mathbf{x}_t)}_{\text{Likelihood score}}
\end{equation}

Using Tweedie's formula \citep{efron2011tweedie}, we approximate the clean data from noisy states:

\begin{equation}
\hat{\mathbf{x}}_0(\mathbf{x}_t) = \mathbb{E}[\mathbf{x}_0 | \mathbf{x}_t] = \mathbf{x}_t + \sigma^2_t s_\theta(\mathbf{x}_t, t)
\end{equation}

This enables approximation of the likelihood score:

\begin{equation}
\nabla_{\mathbf{x}_t}\log p_t(\mathbf{y} | \mathbf{x}_t) \approx -\frac{1}{\sigma_{\text{obs}}^2}\mathbf{H}^T(\mathbf{H}\hat{\mathbf{x}}_0(\mathbf{x}_t) - \mathbf{y})(\mathbf{I} + \sigma_t^2 \nabla_{\mathbf{x}_t} s_\theta(\mathbf{x}_t, t))
\end{equation}

\subsection{Implementation}

The posterior sampling algorithm modifies the reverse diffusion process:

\begin{algorithm}[h]
\caption{Posterior Diffusion Sampling for Hard Data Conditioning}
\label{alg:posterior_sampling_hard_data_appendix}
\KwIn{Observations $\mathbf{y}$, locations $\mathbf{H}$, model $s_\theta$, conditioning strength $\gamma$}
\KwOut{Conditioned permeability field $\mathbf{m}$}
Sample $\mathbf{x}_T \sim \mathcal{N}(0, \sigma_{\text{max}}^2\mathbf{I})$\;
\For{$t$ from $T$ to $0$ with step size $\Delta t$}{
  Compute denoised estimate: $\hat{\mathbf{x}}_0 = \mathbf{x}_t + \sigma_t^2 s_\theta(\mathbf{x}_t, t)$\;
  Compute likelihood gradient: $\mathbf{g}_\text{lik} = -\frac{1}{\sigma_\text{obs}^2}\mathbf{H}^T(\mathbf{H}\hat{\mathbf{x}}_0 - \mathbf{y})$\;
  Update: $\mathbf{x}_{t-\Delta t} = \mathbf{x}_t - \Delta t \cdot g^2(t)[s_\theta(\mathbf{x}_t, t) + \gamma \mathbf{g}_\text{lik}] + g(t)\sqrt{\Delta t} \cdot \mathbf{z}_t$\;
}
Denormalize and apply bounds to obtain $\mathbf{m}$\;
\Return $\mathbf{m}$\;
\end{algorithm}

\subsection{Validation Results}

We validate the approach using three observation scenarios: sparse (grid spacing 16, ~250 points), moderate (grid spacing 8, ~1000 points), and dense (grid spacing 4, ~4000 points). Assessment metrics include matching at observation locations (RMSE < 0.01 in log-space), preservation of channel connectivity between observations, uncertainty quantification in unobserved regions, and geological plausibility of conditioned realizations.

\begin{figure}[htbp]
\centering
\includegraphics[width=0.85\textwidth]{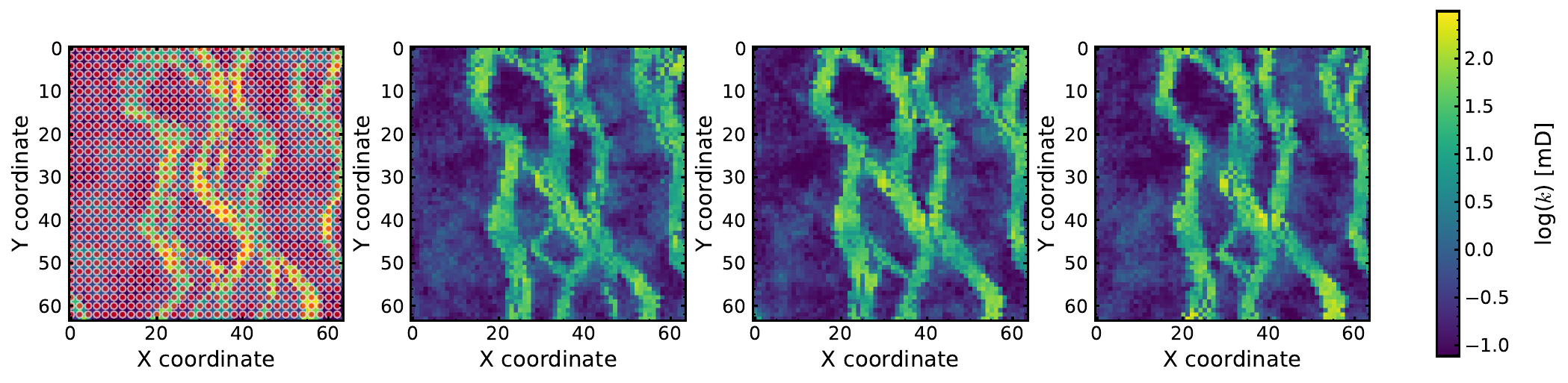}\\[1em]
\includegraphics[width=0.85\textwidth]{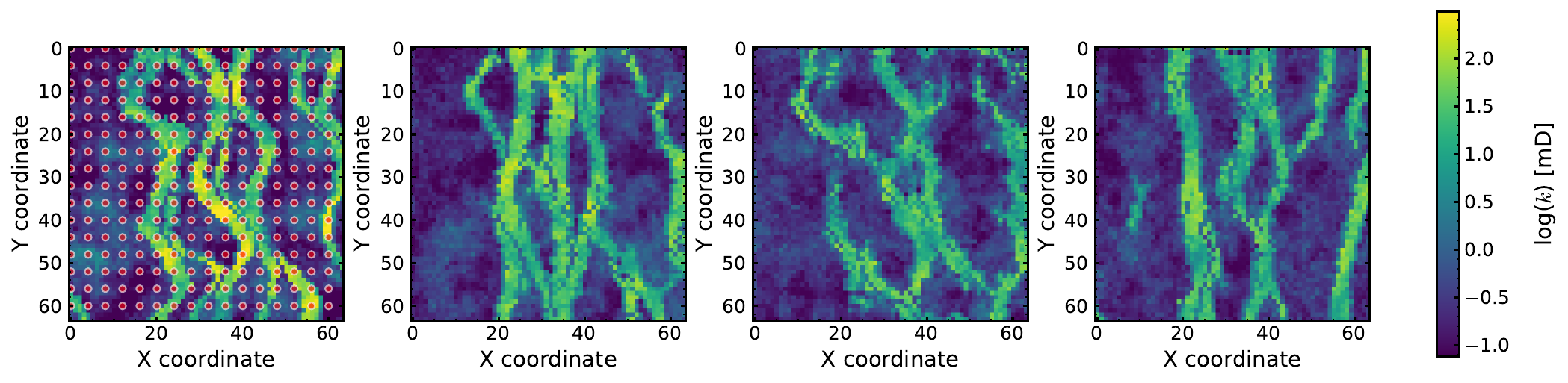}\\[1em]
\includegraphics[width=0.85\textwidth]{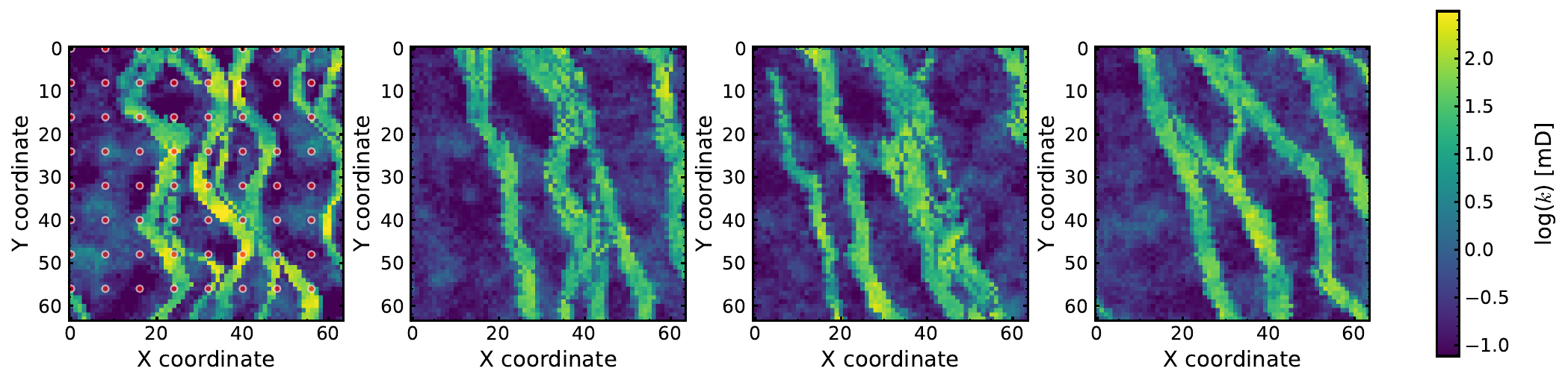}\\[1em]
\includegraphics[width=0.85\textwidth]{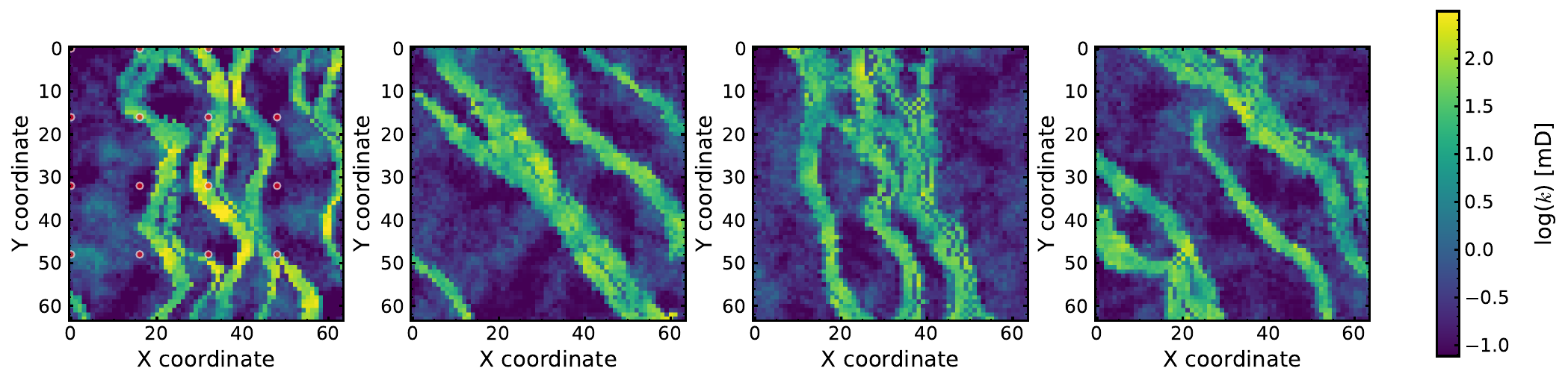}
\caption{Posterior sampling for observation spacings $s = 2, 4, 8, 16$ (approximately 16000, 4000, 1000, 250 points). True fields with observations (white circles) and three conditional realizations per density. Dense observations constrain strongly; sparse observations preserve geological character.}
\label{fig:posterior_samples_multi}
\end{figure}

The uncertainty quantification capabilities are demonstrated through analysis of 100 posterior samples for each observation scenario. The standard deviation maps reveal higher uncertainty in regions distant from observations, as expected. However, unlike pixel-based methods, the uncertainty patterns respect geological continuity, with lower uncertainty along connected channels even at distances from direct observations.

\begin{figure}[htbp]
\centering
\includegraphics[width=\textwidth]{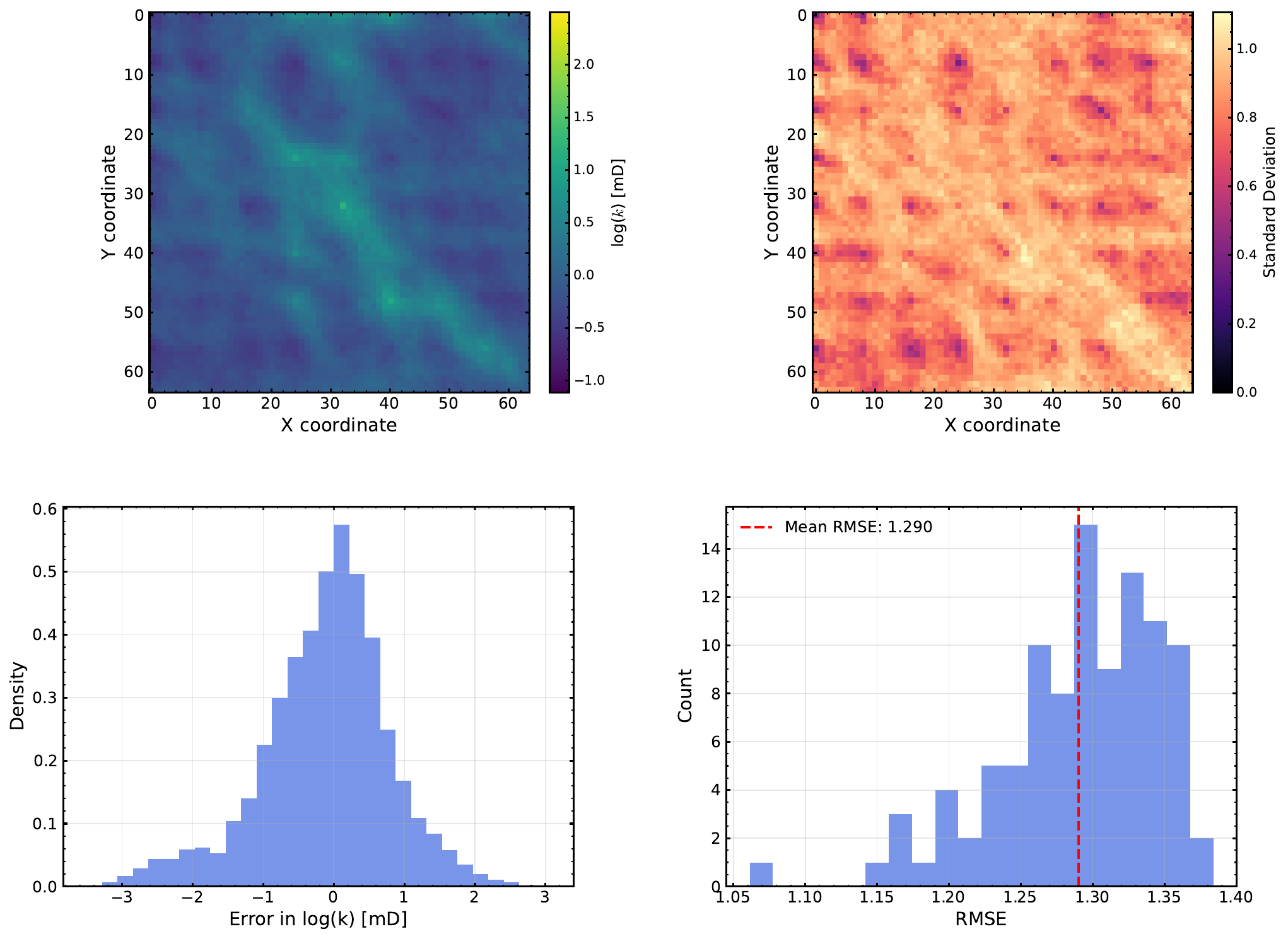}
\caption{Statistical analysis of 100 posterior samples (spacing=8, ~1000 points). (a) Posterior mean, (b) standard deviation following geological features, (c) error histogram confirming matching, (d) preserved variogram structure.}
\label{fig:posterior_statistics_appendix}
\end{figure}

These results demonstrate that the posterior sampling capability provides a tool for integrating hard data constraints while maintaining the geological realism learned by the diffusion model. This approach could be useful in the initial ensemble generation phase when well log data are available, or for creating conditioned realizations for specific scenario analysis in risk assessment studies.

\FloatBarrier

\section{Data Match Plots for Varying Ensemble Sizes}\label{appendix:data_match}
\begin{figure}[H]
\centering
\includegraphics[width=0.7\textwidth]{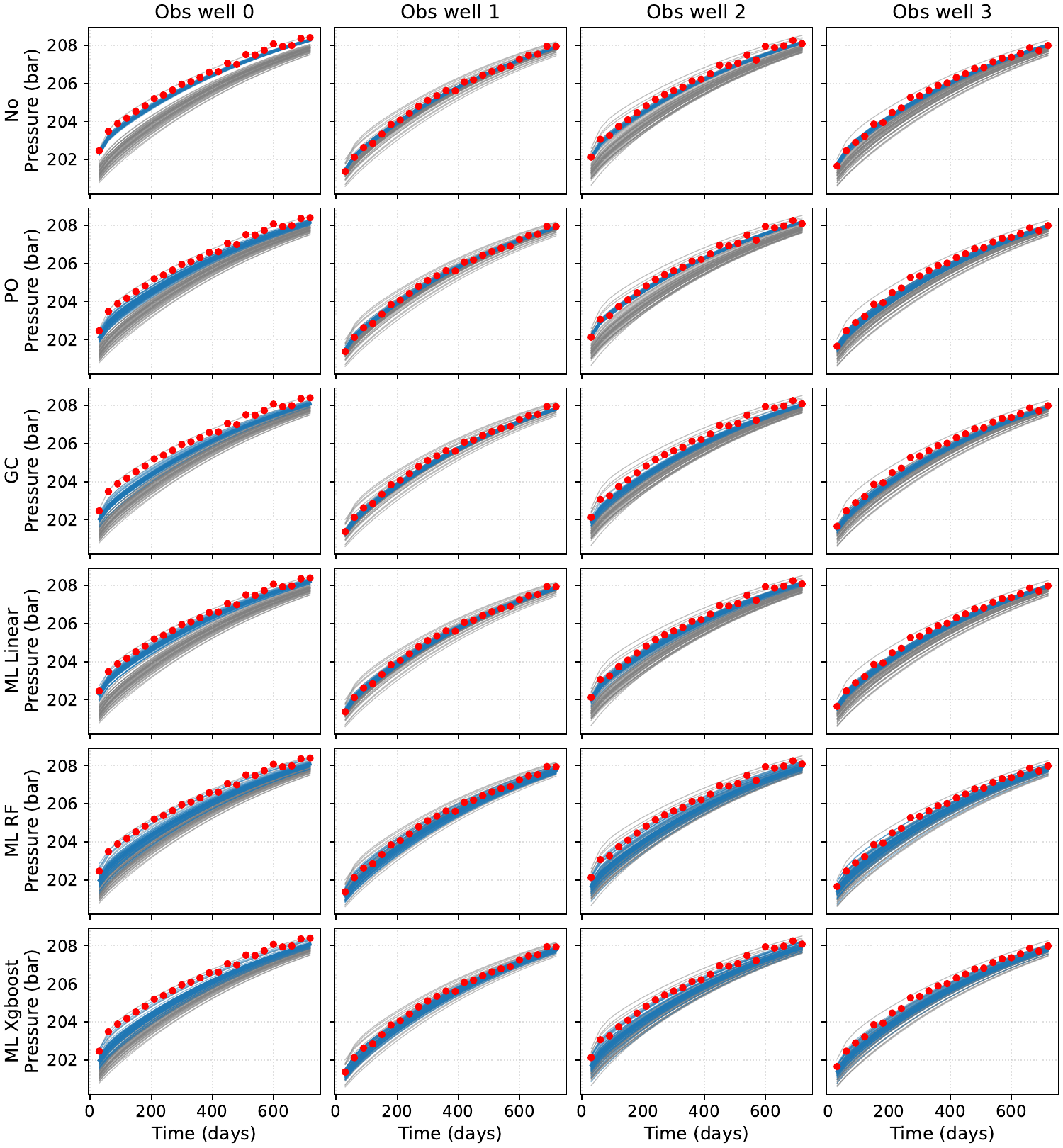}
\caption{Data match results for $N_e = 50$.}
\label{fig:datamatch_ne50}
\end{figure}

\begin{figure}[H]
\centering
\includegraphics[width=0.7\textwidth]{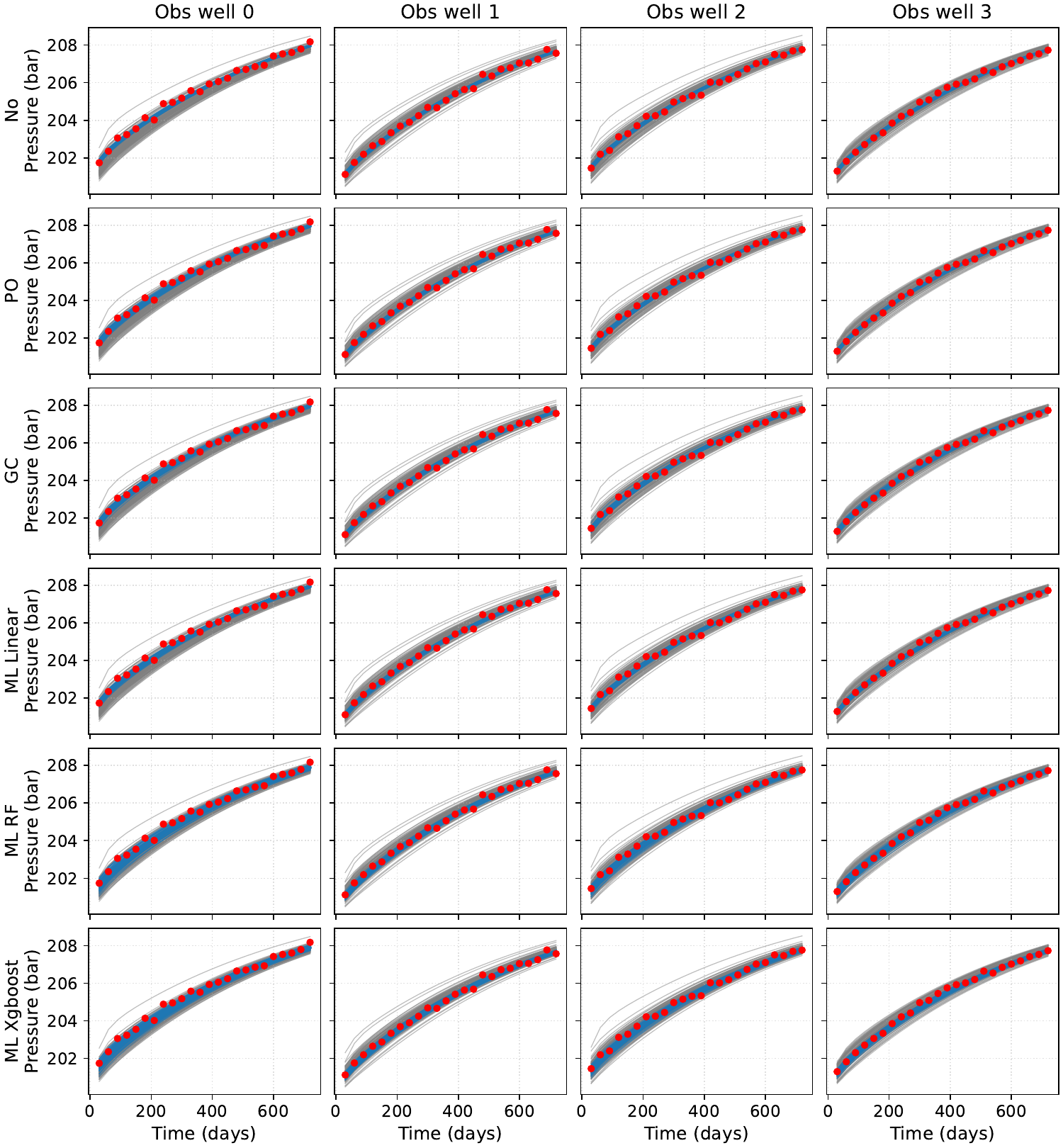}
\caption{Data match results for $N_e = 100$.}
\label{fig:datamatch_ne100}
\end{figure}

\begin{figure}[H]
\centering
\includegraphics[width=0.7\textwidth]{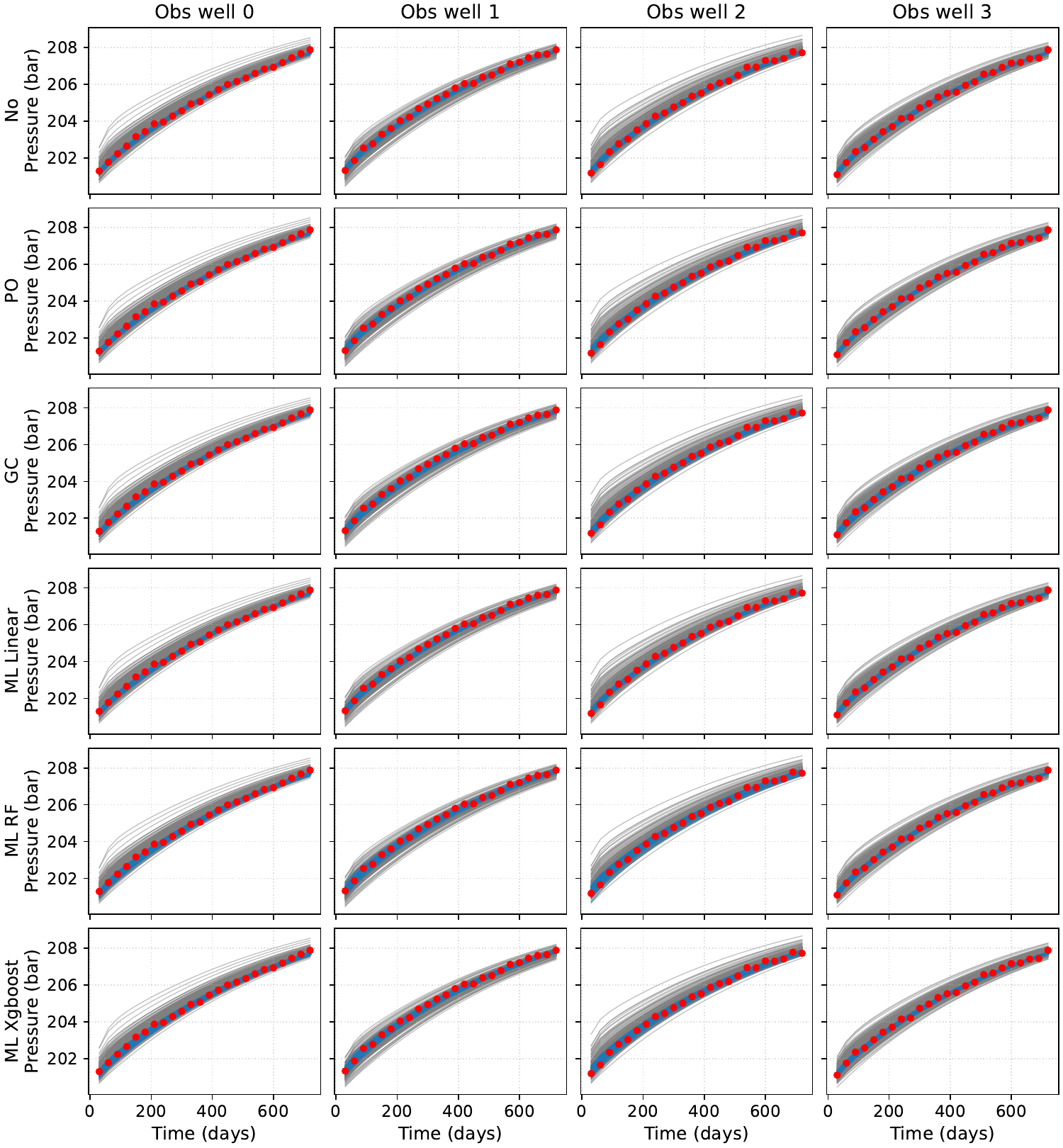}
\caption{Data match results for $N_e = 200$.}
\label{fig:datamatch_ne200}
\end{figure}

\begin{figure}[H]
\centering
\includegraphics[width=0.7\textwidth]{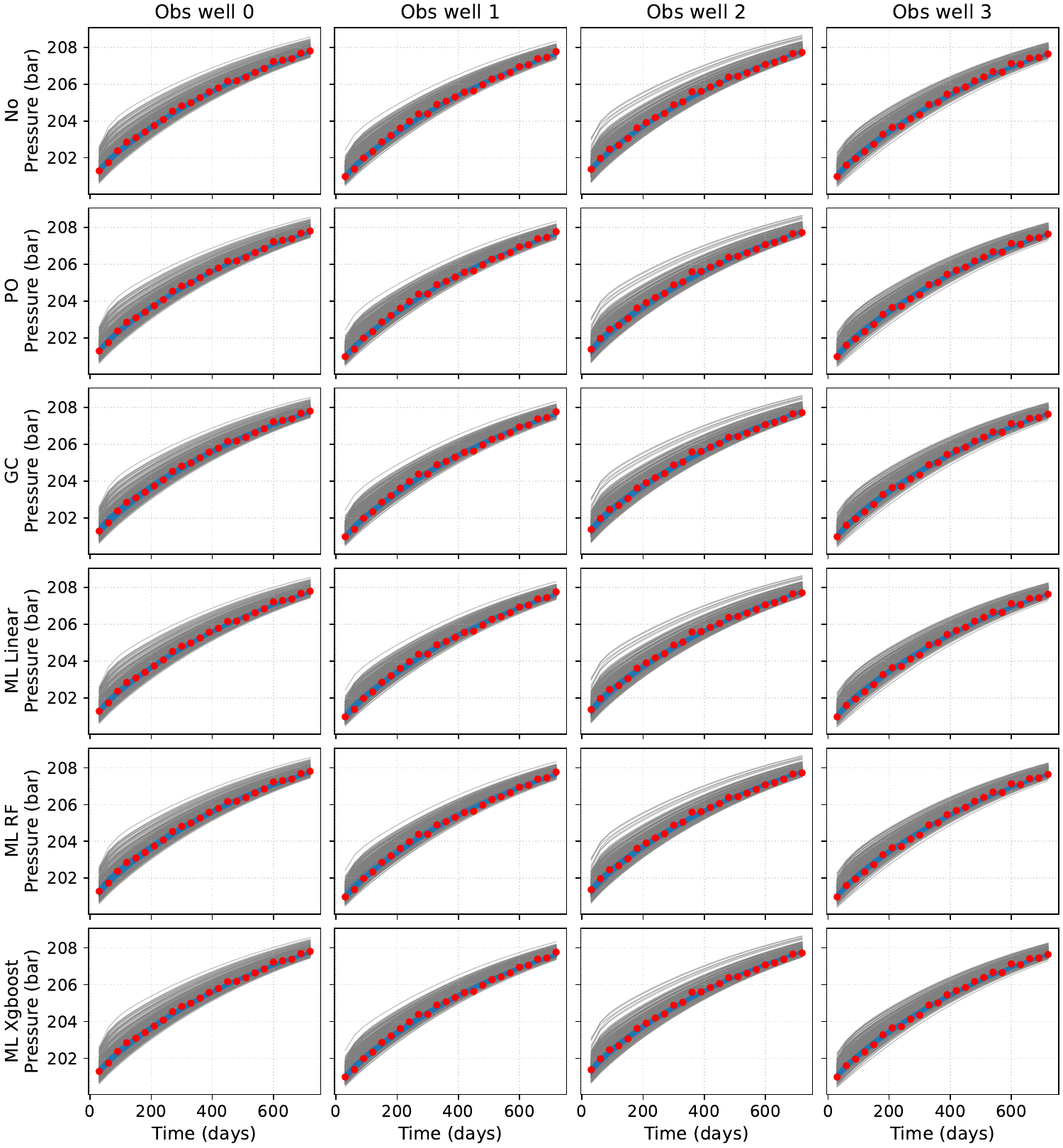}
\caption{Data match results for $N_e = 500$.}
\label{fig:datamatch_ne500}
\end{figure}

\begin{figure}[H]
\centering
\includegraphics[width=0.7\textwidth]{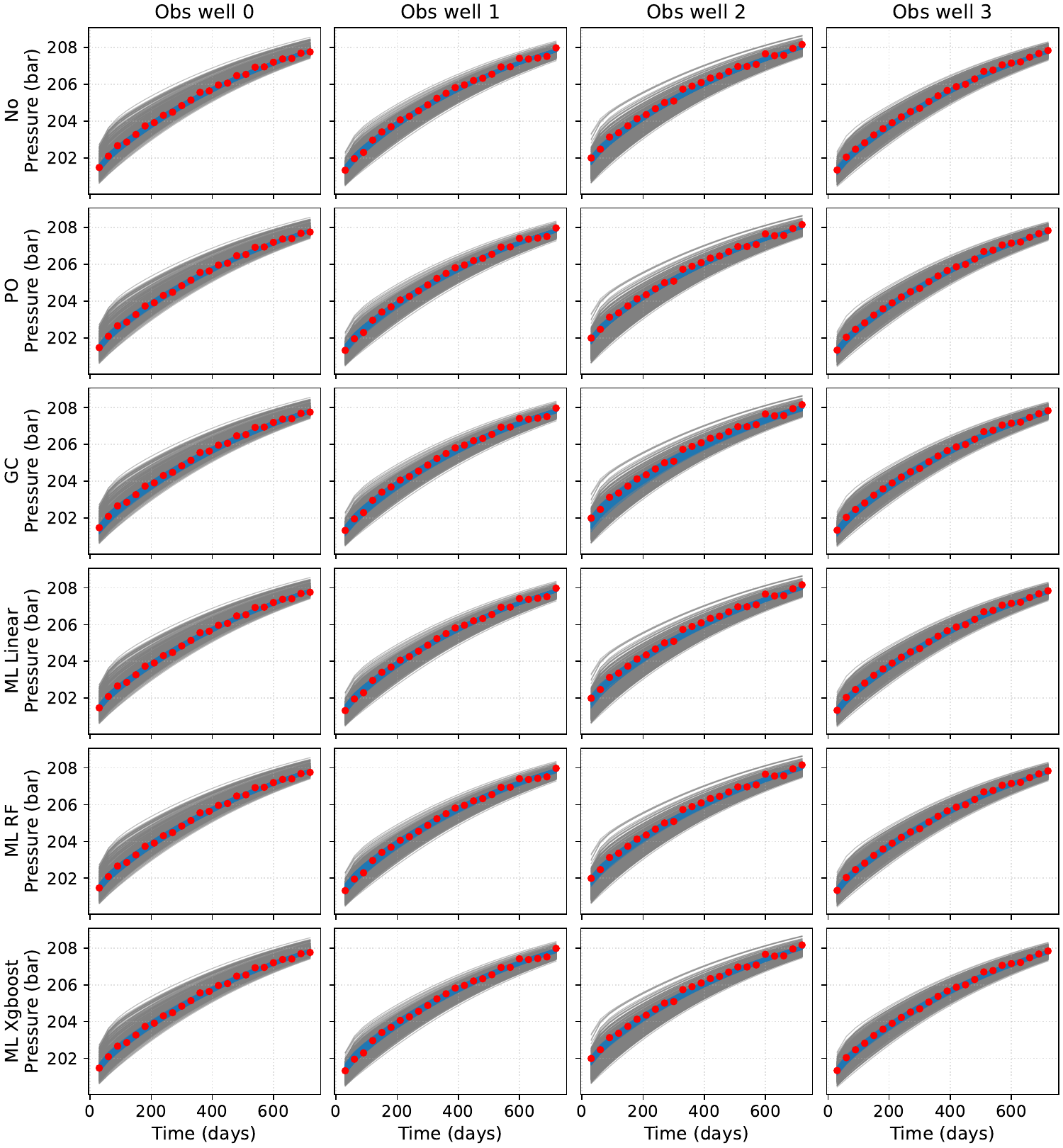}
\caption{Data match results for $N_e = 1000$.}
\label{fig:datamatch_ne999}
\end{figure}

Figures~\ref{fig:datamatch_ne50}--\ref{fig:datamatch_ne999} present the history-matching performance for all ensemble sizes considered in this study. Each figure shows the evolution of pressure at the four monitoring wells over the two-year injection period for the corresponding ensemble size $N_e$.

\section{Localization Maps for Varying Ensemble Sizes}\label{appendix:localization_maps}
This appendix presents the localization maps for all ensemble sizes considered in this study ($N_e = 50, 100, 200, 500, 1000$). Each figure shows the spatial pattern of the localization coefficients for a representative observation well and assimilation time step, illustrating how the localization adapts to the channelized reservoir structure for different ensemble sizes.

\begin{figure}[H]
\centering
\includegraphics[width=0.7\textwidth]{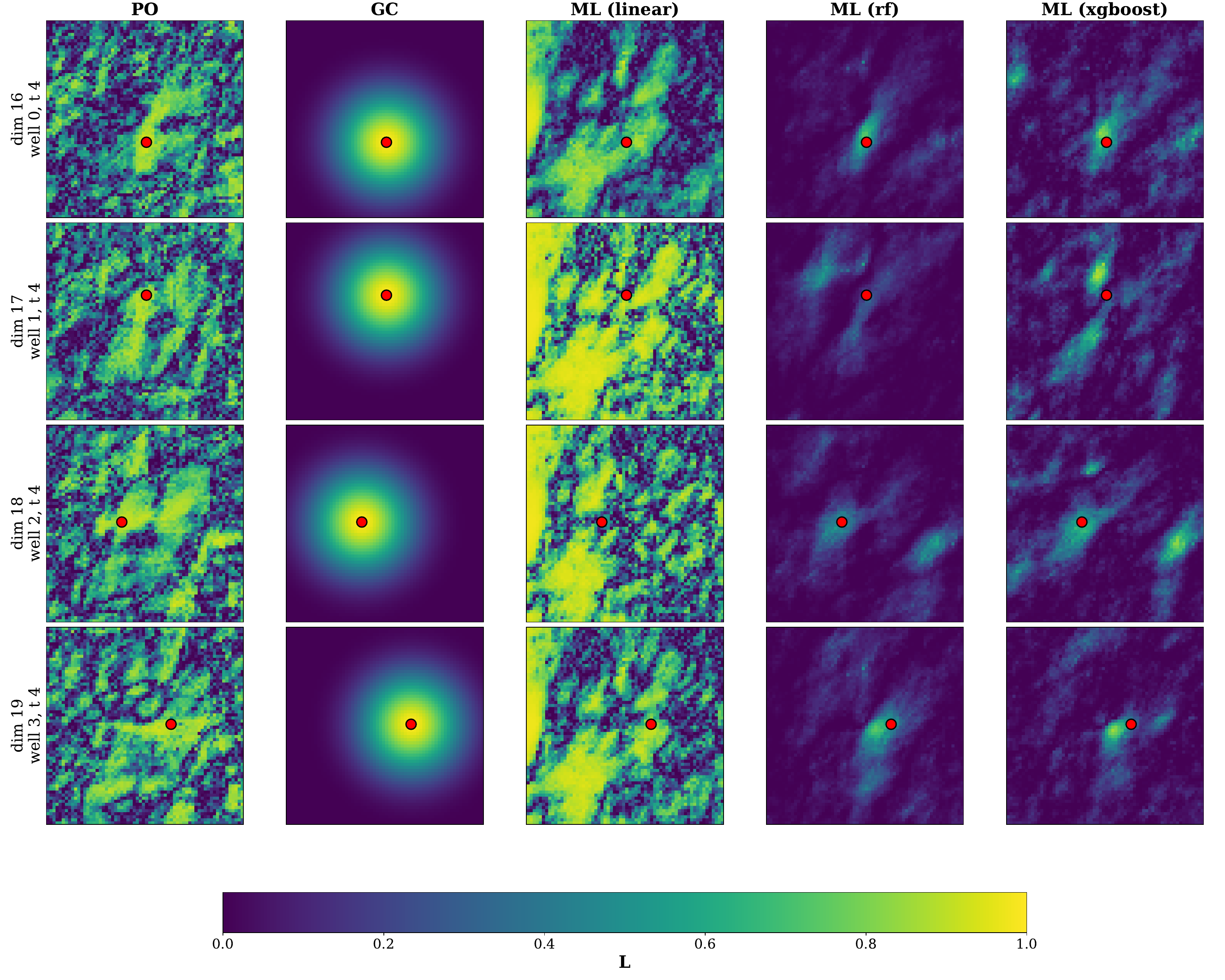}
\caption{Localization map for $N_e = 50$.}
\label{fig:localization_ne50}
\end{figure}

\begin{figure}[H]
\centering
\includegraphics[width=0.7\textwidth]{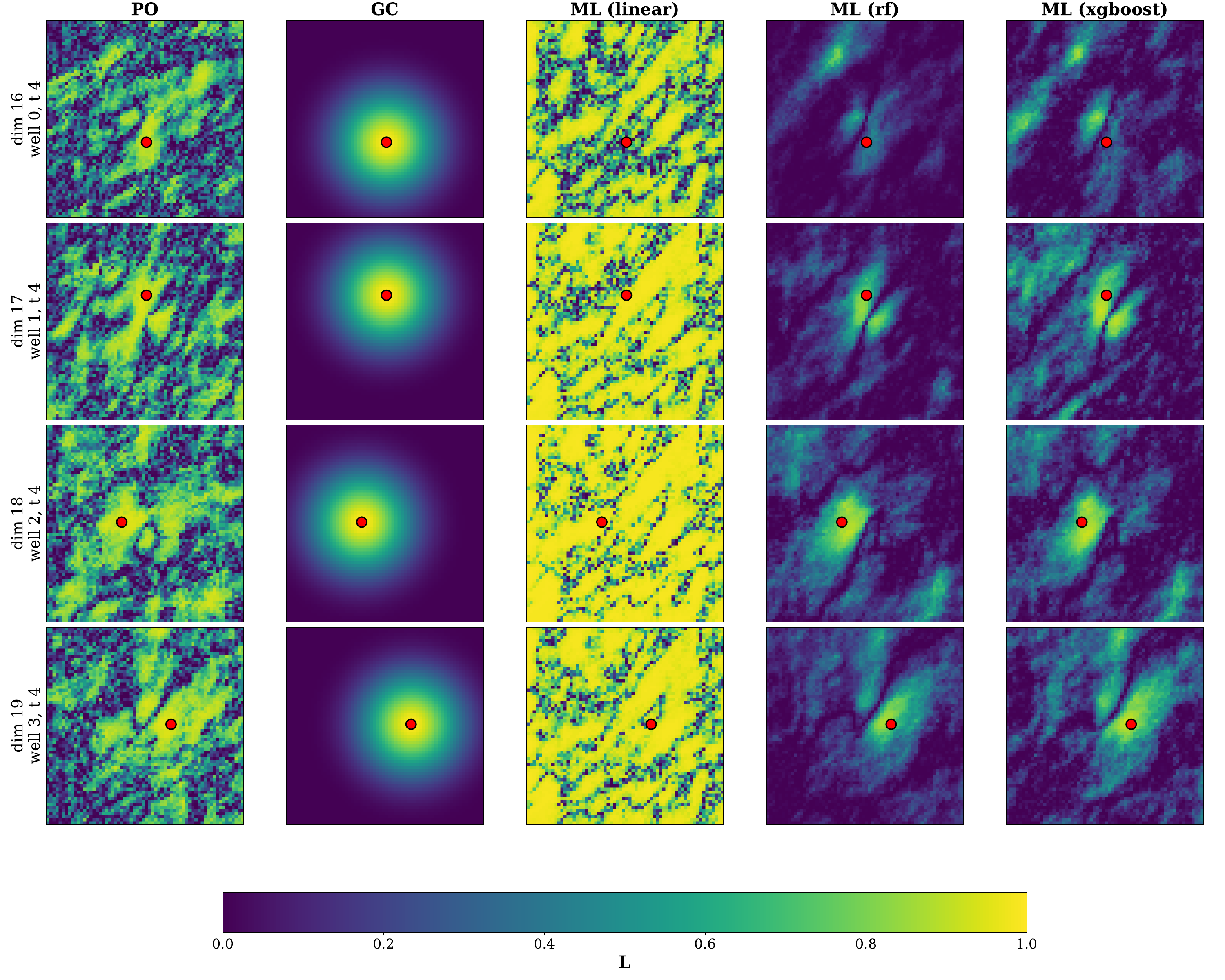}
\caption{Localization map for $N_e = 100$.}
\label{fig:localization_ne100}
\end{figure}

\begin{figure}[H]
\centering
\includegraphics[width=0.7\textwidth]{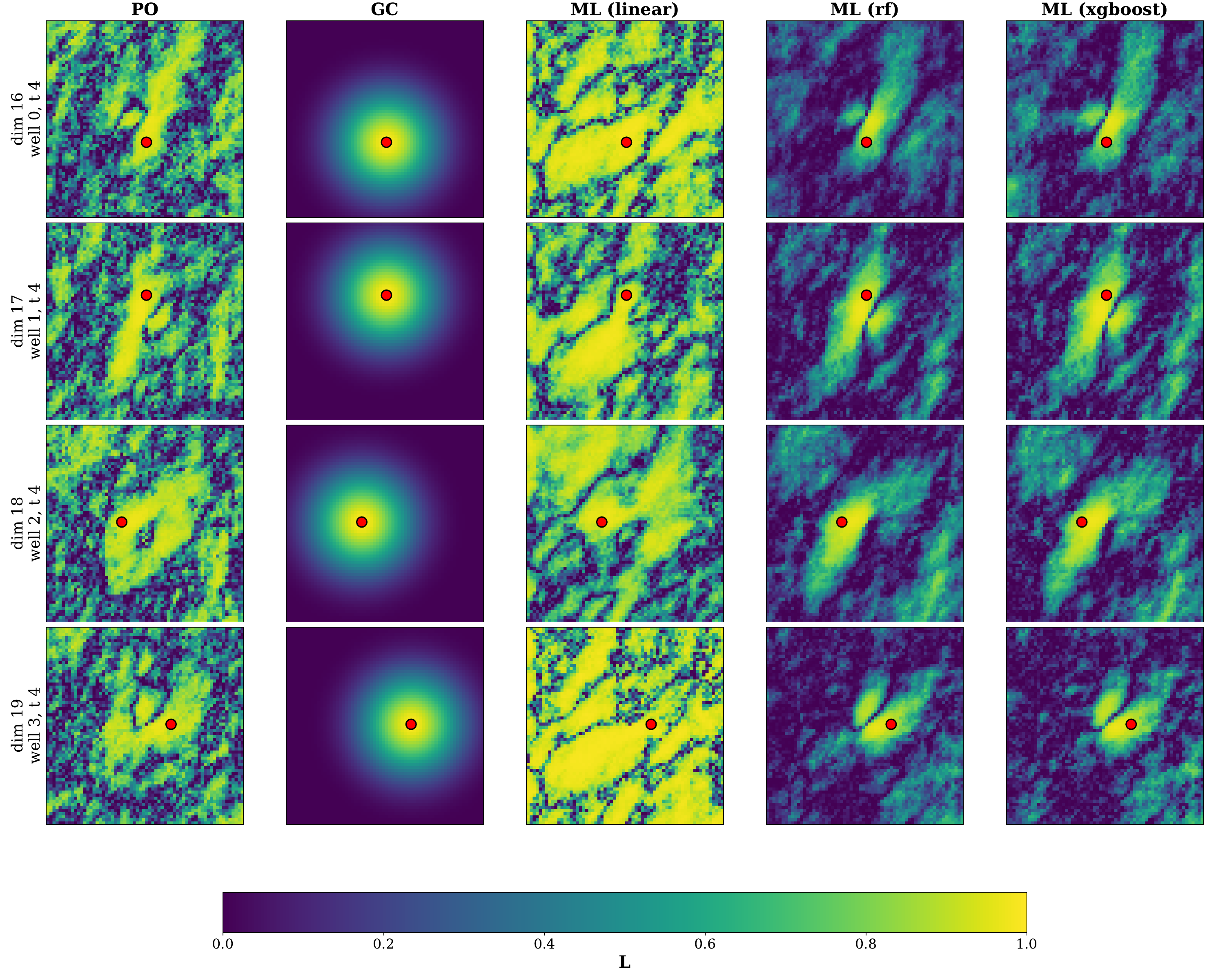}
\caption{Localization map for $N_e = 200$.}
\label{fig:localization_ne200}
\end{figure}

\begin{figure}[H]
\centering
\includegraphics[width=0.7\textwidth]{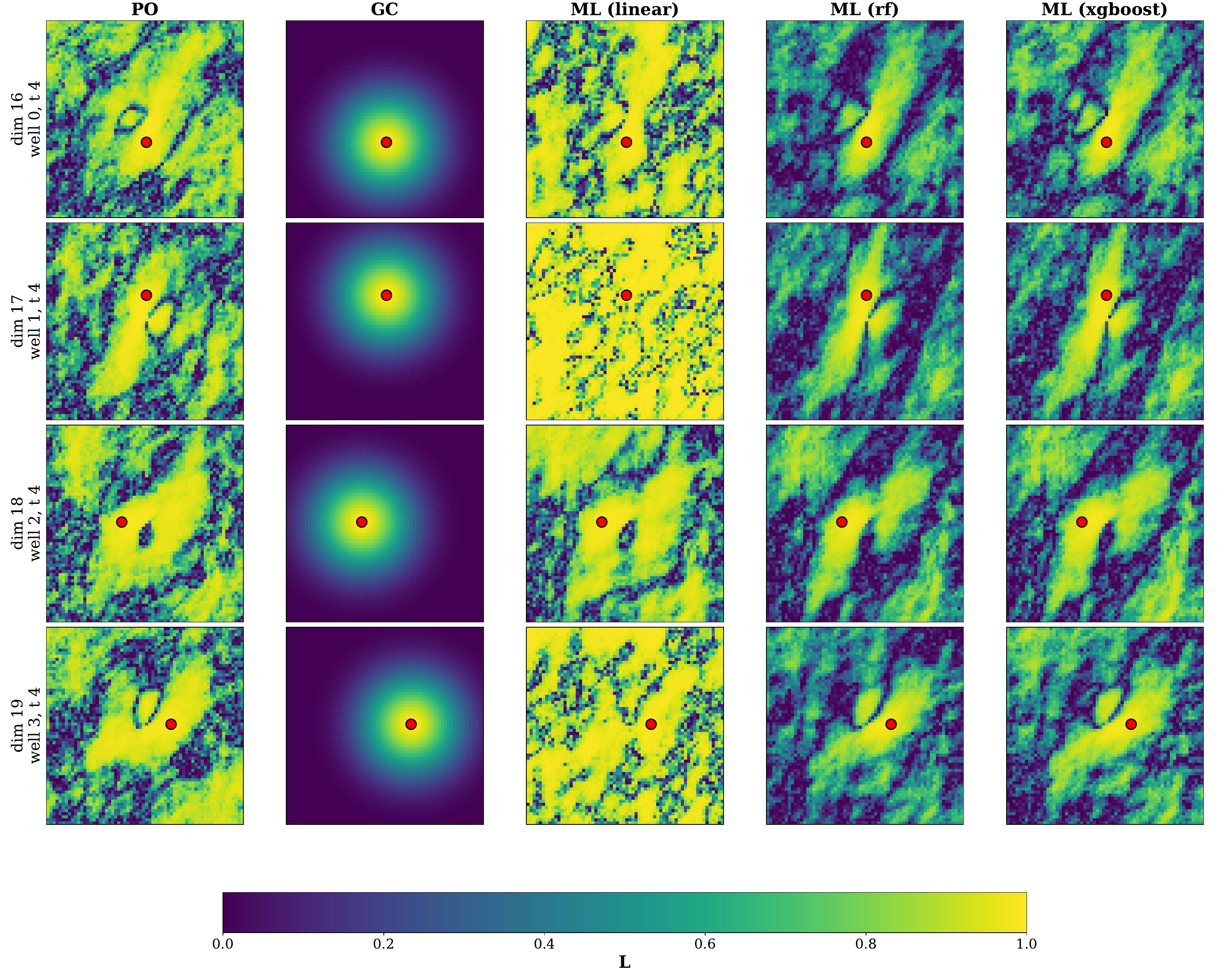}
\caption{Localization map for $N_e = 500$.}
\label{fig:localization_ne500}
\end{figure}

\begin{figure}[H]
\centering
\includegraphics[width=0.7\textwidth]{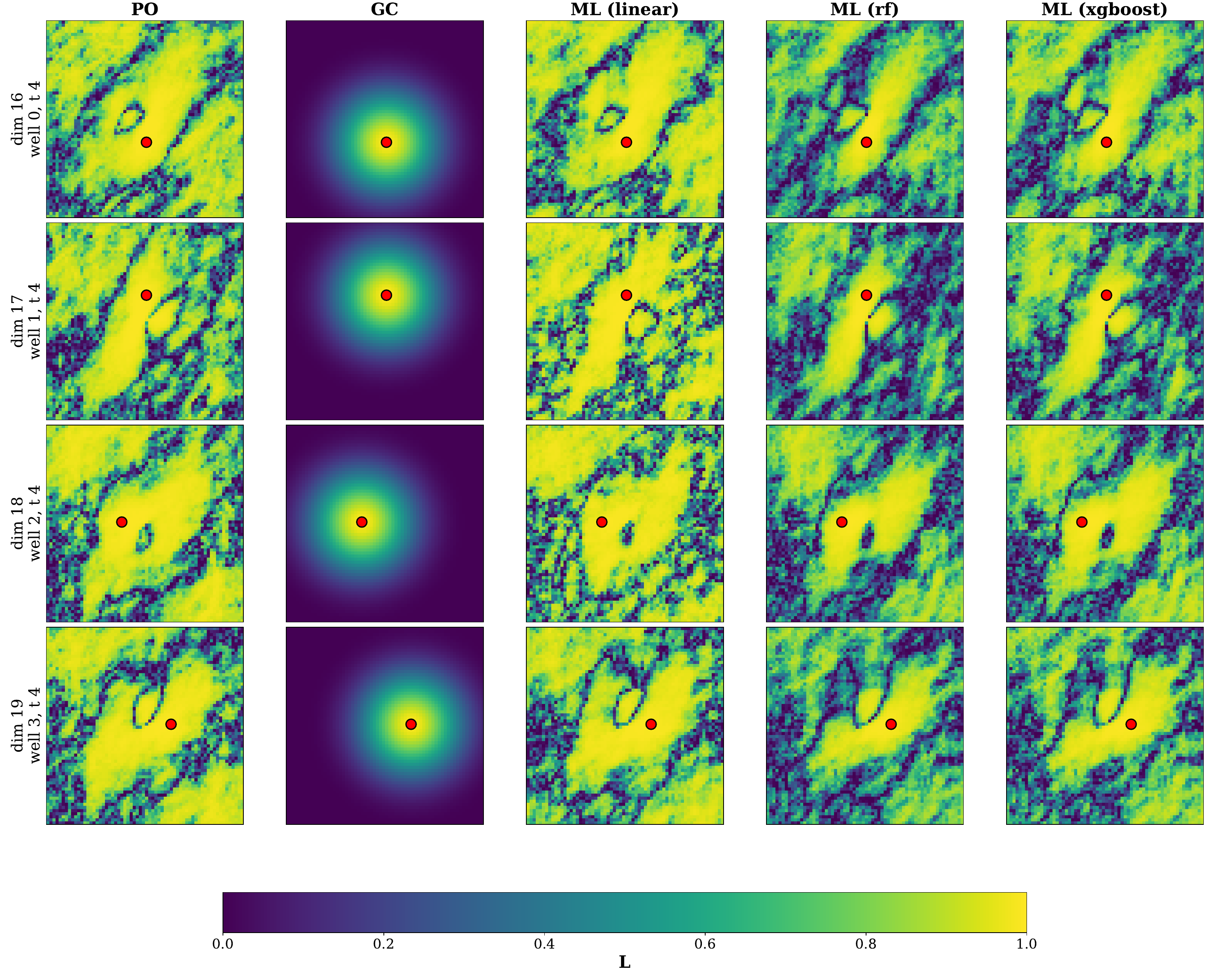}
\caption{Localization map for $N_e = 1000$.}
\label{fig:localization_ne1000}
\end{figure}

\end{document}